\documentclass[10pt,journal,compsoc]{IEEEtran}

\usepackage[nocompress]{cite}
\usepackage[pdftex]{graphicx}
\usepackage{amsmath}
\usepackage{amssymb}
\usepackage{comment}
\usepackage{url}
\usepackage{color}
\usepackage{graphicx}
\usepackage{subfigure} 
\usepackage{subfig}
\usepackage{ragged2e}
\usepackage{booktabs}

\DeclareGraphicsExtensions{.pdf}

\usepackage{xspace}
\makeatletter
\DeclareRobustCommand\onedot{\futurelet\@let@token\@onedot}
\def\@onedot{\ifx\@let@token.\else.\null\fi\xspace}
\def\etal{\emph{et al}\onedot}
\def\ie{\emph{i.e}\onedot} \def\Ie{\emph{I.e}\onedot}

\begin{document}
%
\title{MS-TCN++: Multi-Stage Temporal Convolutional Network for Action Segmentation}
%
%
%
%

\author{Shijie Li$^*$, Yazan Abu Farha$^*$, Yun Liu, Ming-Ming Cheng, Juergen Gall,~\IEEEmembership{Member,~IEEE}
\IEEEcompsocitemizethanks{\IEEEcompsocthanksitem S.~Li, Y.~Abu Farha, and J.~Gall are 
with the University of Bonn, Germany. 
Y.~Liu and M.-M.~Cheng are with the Nankai University, China.\protect\\
S.~Li and Y.~Abu Farha contributed equally.\protect\\
E-mails: \{lishijie, abufarha, gall\}@iai.uni-bonn.de (S.~Li, Y.~Abu Farha, and J.~Gall), vagrantlyun@gmail.com (Y.~Liu), cmm@nankai.edu.cn (M.-M.~Cheng)
}}

\IEEEtitleabstractindextext{%
\begin{abstract}
\justifying
With the success of deep learning in classifying short trimmed videos, more attention has been focused on temporally segmenting and classifying activities in long untrimmed videos.
State-of-the-art approaches for action segmentation utilize several layers of temporal convolution and temporal pooling. 
Despite the capabilities of these approaches in capturing temporal dependencies, their predictions suffer from over-segmentation errors. In this paper, we propose a multi-stage architecture for the temporal action segmentation task that overcomes the limitations of the previous approaches. The first stage generates an initial prediction that is refined by the next ones. In each stage we stack several layers of dilated temporal convolutions covering a large receptive field with few parameters. While this architecture already performs well, lower layers still suffer from a small receptive field. To address this limitation, we propose a dual dilated layer that combines  both large and small receptive fields. We further decouple the design of the first stage from the refining stages to address the different requirements of these stages. Extensive evaluation shows the effectiveness of the proposed model in capturing long-range dependencies and recognizing action segments.
Our models achieve state-of-the-art results on three datasets: 50Salads, Georgia Tech Egocentric Activities (GTEA), and the Breakfast dataset.

\end{abstract}

\begin{IEEEkeywords}
Temporal action segmentation, temporal convolutional network  
\end{IEEEkeywords}}

\maketitle

\IEEEdisplaynontitleabstractindextext

%
\IEEEpeerreviewmaketitle

\IEEEraisesectionheading{\section{Introduction}\label{sec:introduction}}

%
%
%
%
\IEEEPARstart{A}{ction} recognition from video has been an active research area in computer vision in the past few years. 
Most of the efforts, however, have been focused on classifying short trimmed 
videos~\cite{simonyan2014two, feichtenhofer2016spatiotemporal, carreira2017quo, feichtenhofer2019slowfast}. 
Despite the success of these approaches on trimmed videos with a single activity, their performance is limited on long videos containing many action segments. 
Since for many applications, like surveillance and robotics, it is crucial to temporally segment activities in long untrimmed videos, approaches for temporal 
action segmentation have received more attention.
Early attempts for temporal action segmentation tried to extend the success on trimmed videos by combining these models with sliding windows~\cite{rohrbach2012database, karaman2014fast, oneata2014lear}. 
These approaches use temporal windows of different scales to detect and classify action segments. 
However, such approaches are expensive and do not scale for long videos.
Other approaches apply a coarse temporal modeling using Markov models on top of frame-wise classifiers~\cite{kuehne2016end, lea2016segmental, richard2017weakly}. 
While these approaches achieved good results, they are very slow as they require solving a maximization problem over very long sequences.

\begin{figure}[tb]
\begin{center}
\includegraphics[width=.95\linewidth]{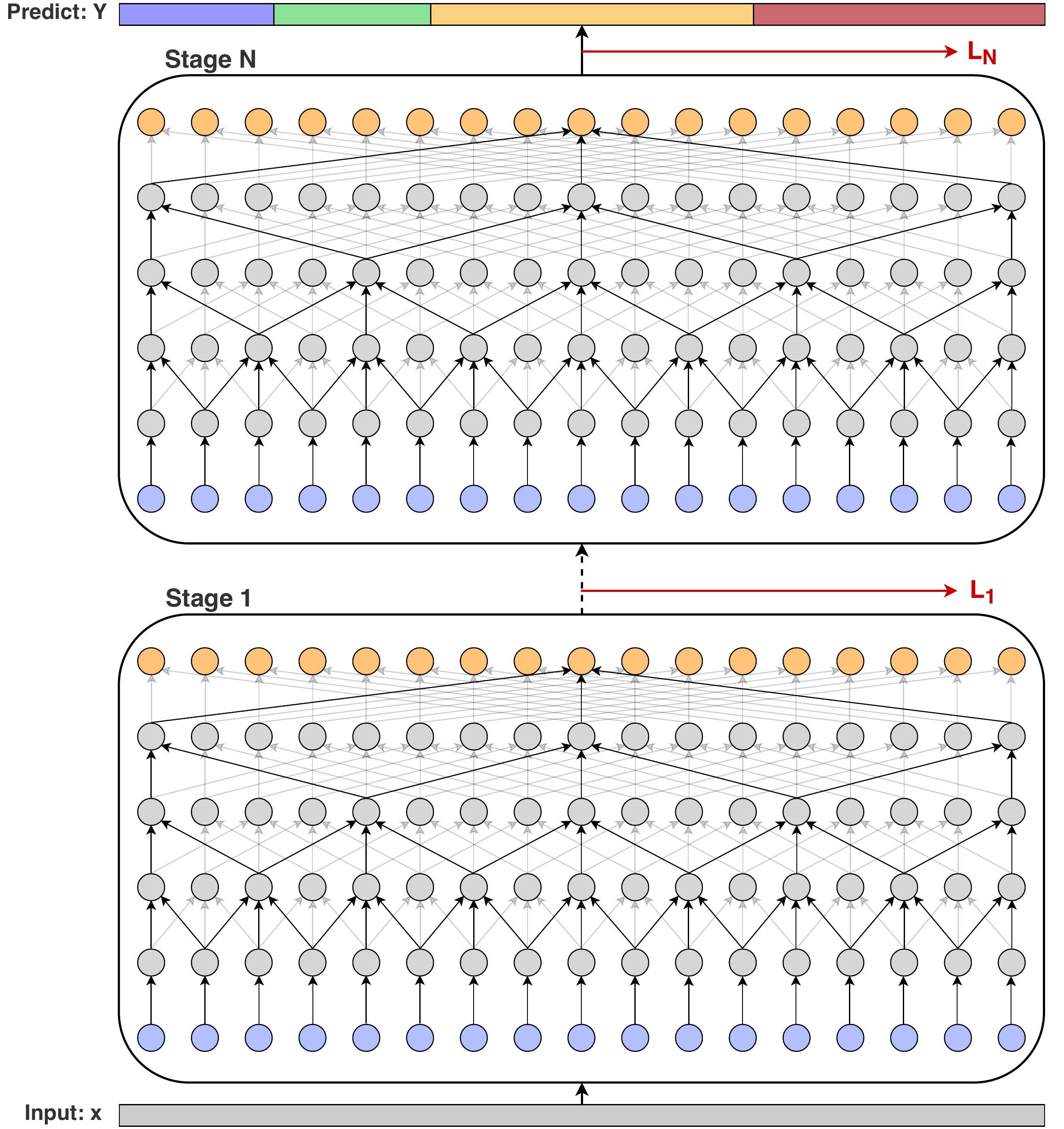}
\end{center}
   \caption{Overview of the multi-stage temporal convolutional network. 
   Each stage generates an initial prediction that is refined by the next stage.
   At each stage, several dilated 1D convolutions are applied on the activations of the previous layer. A loss layer is added after each stage.}
\label{fig:model}
\end{figure}

With the success of temporal convolutional networks (TCNs) as a powerful temporal model for speech synthesis, many researchers adapt TCN-based models for the temporal action segmentation task~\cite{lea2017temporal, lei2018temporal, ding2018weakly}. 
These models were more capable in capturing long range dependencies between the video frames by relying on a large receptive field. 
However, these models are limited for a very low temporal resolution of 
a few frames per second. 
Furthermore, since these approaches rely on temporal pooling layers to increase 
the receptive field, many of the fine-grained information that is necessary for recognition is lost.

To overcome the limitations of the previous approaches, we propose a new model that also uses temporal convolutions. 
In contrast to previous approaches, the proposed model operates on the full temporal resolution of the videos and thus achieves better results. 
Our model consists of multiple stages where each stage outputs an initial prediction that is refined by the next one. 
We call the new architecture Multi-Stage Temporal Convolutional Network (MS-TCN). 
In each stage, we apply a series of dilated 1D convolutions, which enables the model to have a large temporal receptive field with less parameters. Figure~\ref{fig:model} shows an overview of the proposed multi-stage model. 
Furthermore, we employ a smoothing loss during training which penalizes over-segmentation errors in the predictions. 

A preliminary version of this work introducing MS-TCN and the smoothing loss has been published in~\cite{abufarha2019tcn}.
While the proposed MS-TCN already achieves good performance, some of the design choices are sub-optimal. 
First, while the receptive field is very large for higher layers in MS-TCN, lower layers suffer from a small receptive field. 
Second, the first stage in MS-TCN generates an initial prediction and the remaining stages refine this prediction. 
Despite the differences between these two tasks, all stages share the same architecture. 
To address the first limitation, we propose a dual dilated layer (DDL) which combines both large and small 
receptive fields at each layer. For the second, we divide the whole architecture into two parts: 
The first part is the first stage, which is the prediction generation stage, and the second part consists of prediction 
refinement stages. Then we customize the architecture of each part separately and do not force all stages to have 
the same architecture as in MS-TCN. By incorporating these design choices on MS-TCN, we propose an improved version of 
the model, which we call MS-TCN++. Furthermore, we show that the parameters of the refinement stages 
in MS-TCN++ can be shared without compromising the accuracy. This model achieves superior performance compared to MS-TCN 
with much less parameters. Our contribution beyond \cite{abufarha2019tcn} is thus three folded: 
 
\begin{itemize}
    \item We propose a dual dilated layer that combines large and small receptive fields.
    \item We optimize the architecture design of MS-TCN by decoupling the prediction phase and the refinement phase. We call the new model MS-TCN++, which achieves superior results compared to MS-TCN.
    \item We further show that sharing the parameters between the refinement stages in MS-TCN++ results in a more compact model without compromising performance.
\end{itemize}
Extensive evaluation shows the effectiveness of our models in capturing long range dependencies between 
action classes and producing high quality predictions. Our approach achieves state-of-the-art results on 
three challenging benchmarks for action segmentation: 50Salads~\cite{stein2013combining}, Georgia Tech 
Egocentric Activities (GTEA)~\cite{fathi2011learning}, and the Breakfast dataset~\cite{kuehne2014language}. 
Moreover, the proposed models are view-agnostic and work well on all the three datasets, which depict videos with third person view, top view and egocentric videos.

\section{Related Work}
Temporal action segmentation has received a lot of interest from the computer vision 
community. Many approaches where proposed to localize action segments in videos or assign 
action labels to video frames. In earlier approaches, a sliding window 
approach is applied with non-maximum suppression~\cite{rohrbach2012database, karaman2014fast}. 
However, such approaches are computationally expensive since the model has to be evaluated at 
different window scales. Other approaches model actions based on the change in the state of 
objects and materials~\cite{fathi2013modeling} or based on the interactions between hands 
and objects~\cite{fathi2011understanding}. Bhattacharya \etal~\cite{bhattacharya2014recognition} 
use a vector time series representation of videos to model the temporal dynamics of complex 
actions using methods from linear dynamical systems theory. The representation is based on 
the output of pre-trained concept detectors applied on overlapping temporal windows. 
Cheng \etal~\cite{cheng2014temporal} represent videos as a sequence of visual 
words, and model the temporal dependency by employing a Bayesian non-parametric model 
of discrete sequences to jointly classify and segment video sequences.

Despite the success of the previous approaches, their performance was limited as they failed 
in capturing the context over long video sequences. To alleviate this problem, many 
proposals tried to employ high level temporal modeling over frame-wise classifiers.  
Kuehne \etal~\cite{kuehne2016end} represent the frames of a video using Fisher vectors of 
improved dense trajectories, and then each action is modeled with a hidden Markov 
model (HMM). These HMMs are combined with a context-free grammar for recognition 
to determine the most probable sequence of actions. HMMs are also used in many other approaches. 
\cite{kuehne2017weakly} combine HMMs with a Gaussian mixture model (GMM) as a frame-wise classifier. 
However, since frame-wise classifiers do not capture enough context to detect action classes, 
Richard \etal~\cite{richard2017weakly} and Kuehne \etal~\cite{kuehne2018hybrid} use a GRU instead of the 
GMM that is used in~\cite{kuehne2017weakly}. A hidden Markov model is also 
used in~\cite{tang2012learning} to model both transitions between states and their durations.
Vo and Bobick~\cite{vo2014stochastic} use a Bayes network to segment activities. 
They represent compositions of actions using a stochastic context-free grammar with 
AND-OR operations.
\cite{richard2016temporal} propose a model for temporal action detection that 
consists of three components:  an action model that maps features extracted from 
the video frames into action probabilities, a language model that describes the 
probability of actions at sequence level, and finally a length model that models 
the length of different action segments. To get the video segmentation, they use 
dynamic programming to find the solution that maximizes the joint probability of 
the three models.
Singh \etal~\cite{singh2016multi} use a two-stream network to learn representations 
of short video chunks. These representations are then passed to a bi-directional 
LSTM to capture dependencies between different chunks. However, their approach is 
very slow due to the sequential prediction.
In~\cite{singh2016first}, a three-stream architecture that operates on spatial, 
temporal and egocentric streams is introduced to learn  egocentric-specific features. 
These features are then classified using a multi-class SVM.

Multi-scale information is important for speech synthesis~\cite{van2016wavenet} and image recognition~\cite{gao2019res2net}.
Inspired by these,
researchers have tried to use similar ideas for the temporal action segmentation 
task. Lea \etal~\cite{lea2017temporal} propose a temporal convolutional network for action 
segmentation and detection. Their approach follows an encoder-decoder architecture with 
a temporal convolution and pooling in the encoder, and upsampling followed by deconvolution 
in the decoder. While using temporal pooling enables the model to capture long-range dependencies, 
it might result in a loss of fine-grained information that is necessary for fine-grained recognition. 
Lei and Todorovic~\cite{lei2018temporal} build on top of~\cite{lea2017temporal} and use deformable 
convolutions instead of the normal convolution and add a residual stream to the encoder-decoder model. 
Ding and Xu~\cite{ding2018weakly} add lateral connections to the encoder-decoder TCN~\cite{lea2017temporal} 
and propose a temporal convolutional feature pyramid network for predicting frame-wise action labels.
All of the approaches in~\cite{lea2017temporal, lei2018temporal, ding2018weakly} operate on downsampled 
videos with a temporal resolution of 1-3 frames per second. 
In contrast to these approaches, we operate on the full temporal resolution and 
use dilated convolutions to capture long-range dependencies. 
Recently, Mac \etal~\cite{mac2019learning} propose to learn spatio-temporal features using deformable 
convolutions and local consistency constraints. On the contrary, in our approach we only focus on 
the long-term temporal modeling.

Action detection is a related but a different task. In this context, the goal is to detect sparse 
action segments while most parts of the videos are unlabeled. In this work, we focus on action 
segmentation where the videos are densely annotated. For action detection, several approaches follow 
a two stage pipeline. 
The first stage is to generate proposals, and then classify and refine the boundaries of these proposals 
in the second stage~\cite{shou2016temporal, xu2017r, zhao2017temporal, chao2018rethinking, zeng2019graph, gao2017turn, buch2017sst, gao2017cascaded, lin2019bmn}. 
Other approaches combine the proposal generation and classification in a single-stage architecture which 
enables end-to-end training~\cite{yeung2016end, buch2017end, long2019gaussian}.

\section{Temporal Action Segmentation}
We introduce a multi-stage temporal convolutional network (MS-TCN) for the temporal action segmentation task. 
Then we introduce a new layer and address the limitations of MS-TCN to propose an improved model called MS-TCN++.
Given the frames of a video $x_{1:T} = (x_1,\dots,x_T)$, our goal is to infer the class label for each 
frame $c_{1:T} = (c_1,\dots,c_T)$, where $T$ is the video length. %
First, we describe the single-stage approach in Section~\ref{sec:single_stage_model}, then we discuss the 
multi-stage model in Section~\ref{sec:multi_stage_model}. Section~\ref{sec:ddl} introduces the dual dilated layer. 
In Section~\ref{sec:MS-TCN++}, we analyze the drawbacks of MS-TCN and introduce the improved model MS-TCN++. 
Finally, we describe the proposed loss function in Section~\ref{sec:loss_function}.

\begin{figure}[tb]
\begin{center}
   \includegraphics[width=0.47\linewidth]{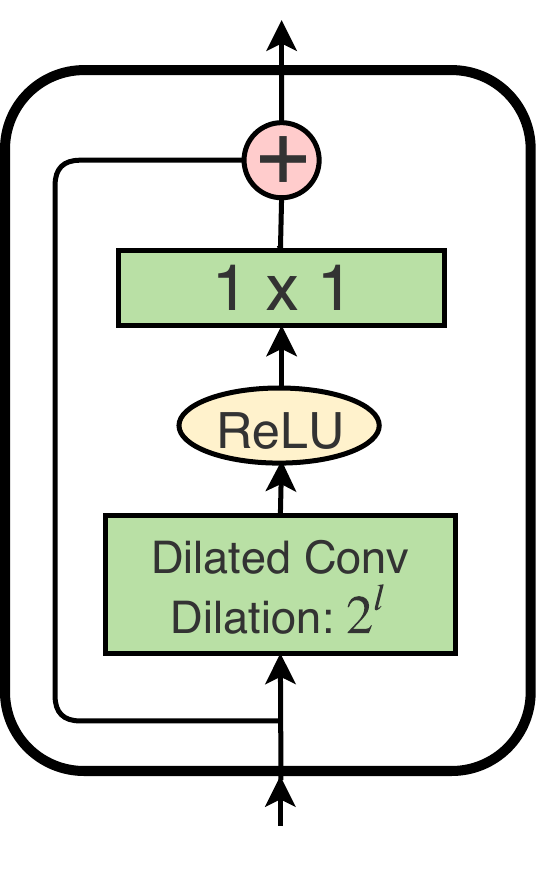}
\end{center}
   \caption{Overview of the dilated residual layer. At each layer $l$, the dilated residual layer uses 
   a convolution with dilated factor $2^l$.}
\label{fig:layer}
\end{figure}

\subsection{Single-Stage TCN}
\label{sec:single_stage_model}
Our single stage model consists of only temporal convolutional layers. 
We do not use pooling layers, which reduce the temporal resolution, or 
fully connected layers, which force the model to operate on inputs of 
fixed size and massively increase the number of parameters. We call this 
model a single-stage temporal convolutional network (SS-TCN).
The first layer of a single-stage TCN is a $1 \times 1$ convolutional layer, 
that adjusts the dimension of the input features to match the number 
of feature maps in the network. Then, this layer is followed by several 
layers of dilated 1D convolution. Inspired by the wavenet~\cite{van2016wavenet} 
architecture, we use a dilation factor that is doubled at each layer, 
\ie $1, 2, 4, ...., 512$. All these layers have the same number of 
convolutional filters. However, instead of the causal convolution that 
is used in wavenet, we use acausal convolutions with kernel size 3. 
Each layer applies a dilated convolution with ReLU activation to the output 
of the previous layer. We further use residual connections to facilitate 
gradients flow. The set of operations at each layer can be formally 
described as follows
 \begin{align}
& \hat{H}_l = ReLU(W_d * H_{l-1} + b_d), \\
& H_l = H_{l-1} + W * \hat{H}_l + b, 
\end{align}
where $H_l$ is the output of layer $l$, $*$ denotes the convolution operator, 
$W_d \in \mathbb{R}^{3 \times D \times D}$ are the weights of the dilated convolution 
filters with kernel size 3 and $D$ is the number of convolutional filters, 
$W \in \mathbb{R}^{1 \times D \times D}$ are the weights of a $1 \times 1$ convolution, 
and $b_d, b \in \mathbb{R}^{D}$ are bias vectors. These operations are illustrated in 
Figure~\ref{fig:layer}. Using dilated convolution increases the receptive field 
without the need to increase the number of parameters by increasing the number 
of layers or the kernel size. Since the receptive field grows exponentially with 
the number of layers, we can achieve a very large receptive field with a few layers, 
which helps in preventing the model from over-fitting the training data. The receptive 
field at each layer is determined by 
\begin{equation}
ReceptiveField(l) = 2^{l+1} - 1, 
\end{equation}
where $l \in \left[ 1, L\right] $ is the layer number. Note that this formula is only 
valid for a kernel of size 3. To get the probabilities for the output class, we apply a 
$1 \times 1$ convolution over the output of the last dilated convolution layer followed 
by a softmax activation, \ie
\begin{equation}
Y_t = Softmax(Wh_{L,t} + b), 
\end{equation}
where $Y_t$ contains the class probabilities at time $t$, $h_{L,t}$ is the output 
of the last dilated convolution layer at time $t$, $W \in \mathbb{R}^{C \times D}$ and 
$b \in \mathbb{R}^{C}$ are the weights and bias for the $1 \times 1$ convolution layer. 
$C$ is the number of classes and $D$ is the number of convolutional filters.

\subsection{Multi-Stage TCN}
\label{sec:multi_stage_model}
Stacking several predictors sequentially has shown significant 
improvements in many tasks like human pose estimation~\cite{wei2016convolutional, 
newell2016stacked, dantone2014body}. The idea of these stacked or multi-stage architectures 
is composing several models sequentially such that each model operates directly on 
the output of the previous one. The effect of such composition is 
an incremental refinement of the predictions from the previous stages. 

Motivated by the success of such architectures, we introduce a multi-stage 
temporal convolutional network (MS-TCN) for the temporal action segmentation
task. In this multi-stage model, each stage takes an initial prediction from 
the previous stage and refines it. The input of the first stage are the 
frame-wise features of the video as follows
\begin{align}
& Y^0 = x_{1:T}, \\
& Y^s = \mathcal{F}(Y^{s-1}), 
\end{align}
where $Y^s$ is the output at stage $s$ and $\mathcal{F}$ is the single-stage TCN 
discussed in Section~\ref{sec:single_stage_model}. Using such a multi-stage 
architecture helps in providing more context to predict the class label at each 
frame. Furthermore, since the output of each stage is an initial prediction, 
the network is able to capture dependencies between action classes and learn 
plausible action sequences, which helps in reducing the over-segmentation errors.

Note that the input to the next stage is just the frame-wise probabilities without 
any additional features. We will show in the experiments how adding features to the 
input of the next stage affects the quality of the predictions.

\subsection{Dual Dilated Layer }
\label{sec:ddl}
\begin{figure}[tb]
\begin{center}
\vspace{1mm}
\includegraphics[width=.9\linewidth]{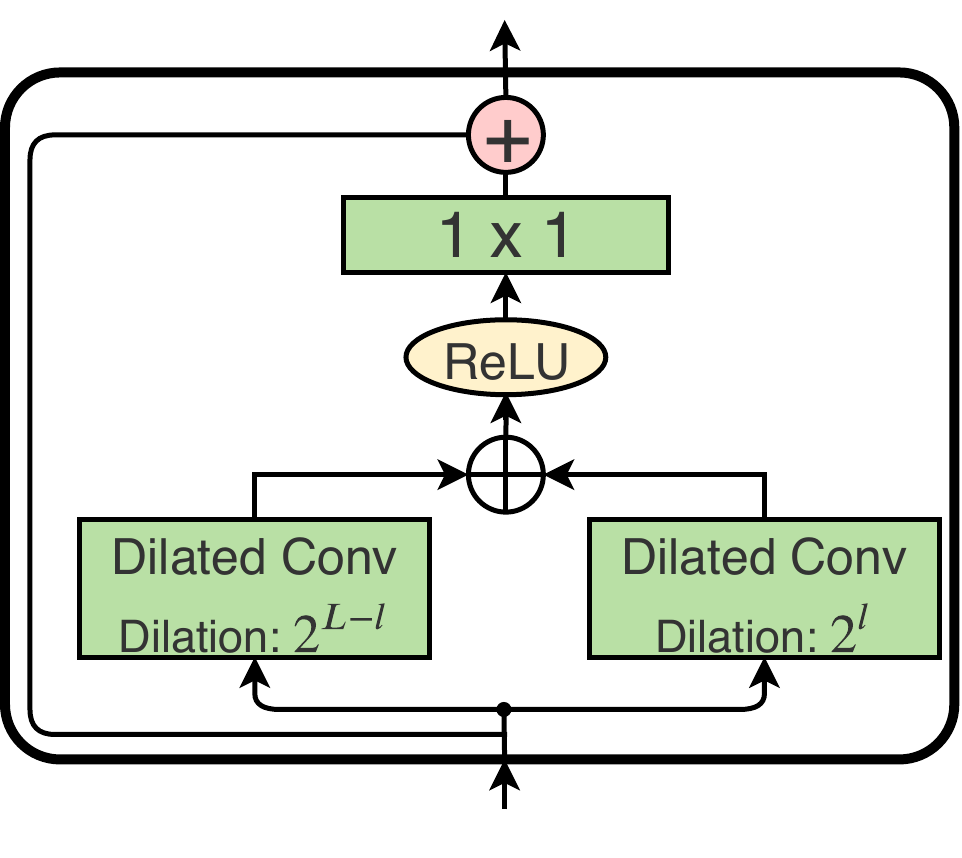}
\end{center}
   \caption{Overview of the dual dilated layer (DDL). At each layer $l$, DDL uses two convolutions with dilated factor $2^l$ and $2^{L-l}$, respectively, where $L$ is the number of layers in the network.} 
   \label{fig:ddl}
\end{figure}

In the dilated convolution layers in MS-TCN, the dilation factor increases as we increase the number 
of layers. While this results in a large receptive field for higher layers, lower layers still suffer 
from very low receptive fields. Furthermore, higher layers in MS-TCN apply convolutions over very distant 
time steps due to the large dilation factor. To overcome this problem, we propose a dual dilated layer 
(DDL). Instead of having one dilated convolution, the DDL combines two convolutions with different dilation factor. 
The first convolution has a low dilation factor in lower layers and exponentially increases as we increases 
the number of layers. Whereas for the second convolution, we start with a large dilation factor in lower 
layers and exponentially decrease it with increasing the number of layers. The set of operations at each 
layer can be formally described as follows
\begin{align}
& \hat{H}_{l,d_1} = W_{d_1} * H_{l-1} + b_{d_1}, \\
& \hat{H}_{l,d_2} = W_{d_2} * H_{l-1} + b_{d_2}, \\
& \hat{H}_l = ReLU([\hat{H}_{l,d_1}, \hat{H}_{l,d_2}]), \label{eq:concat}\\
& H_l = H_{l-1} + W * \hat{H}_l + b, 
\end{align}
where $W_{d_1}, W_{d_2} \in \mathbb{R}^{3 \times D \times D}$ are the weights of dilated convolutions with 
dilation factor $2^l$ and $2^{L-l}$ respectively, $W \in \mathbb{R}^{1 \times 2D \times D}$ are the weights of 
a $1 \times 1$ convolution, and $b_{d_1}, b_{d_2}, b \in \mathbb{R}^{D}$ are bias vectors. In (\ref{eq:concat}), 
$\hat{H}_{l,d_1}$ and $ \hat{H}_{l,d_2}$ are concatenated. An overview of the dual dilated layer 
is illustrated in Figure~\ref{fig:ddl}.

While the dual dilated layer combines local and global context from the input sequence, there are other techniques in the literature for fusing multi-scale features like feature 
pyramid networks (FPN)~\cite{lin2017feature}. While applying FPN for temporal action segmentation has been 
successful~\cite{ding2018weakly}, these approaches still suffer from a very limited receptive field. Moreover, 
the multi-scale features in FPN are obtained by applying pooling operations which results in a loss of the 
fine-grained information that is necessary for temporal segmentation. On the contrary, DDL combines multi-scale features 
and yet preserves the temporal resolution of the input sequence.

\subsection{MS-TCN++}
\label{sec:MS-TCN++}
\begin{figure}[tb]
\vspace{1mm}
\includegraphics[width=\linewidth]{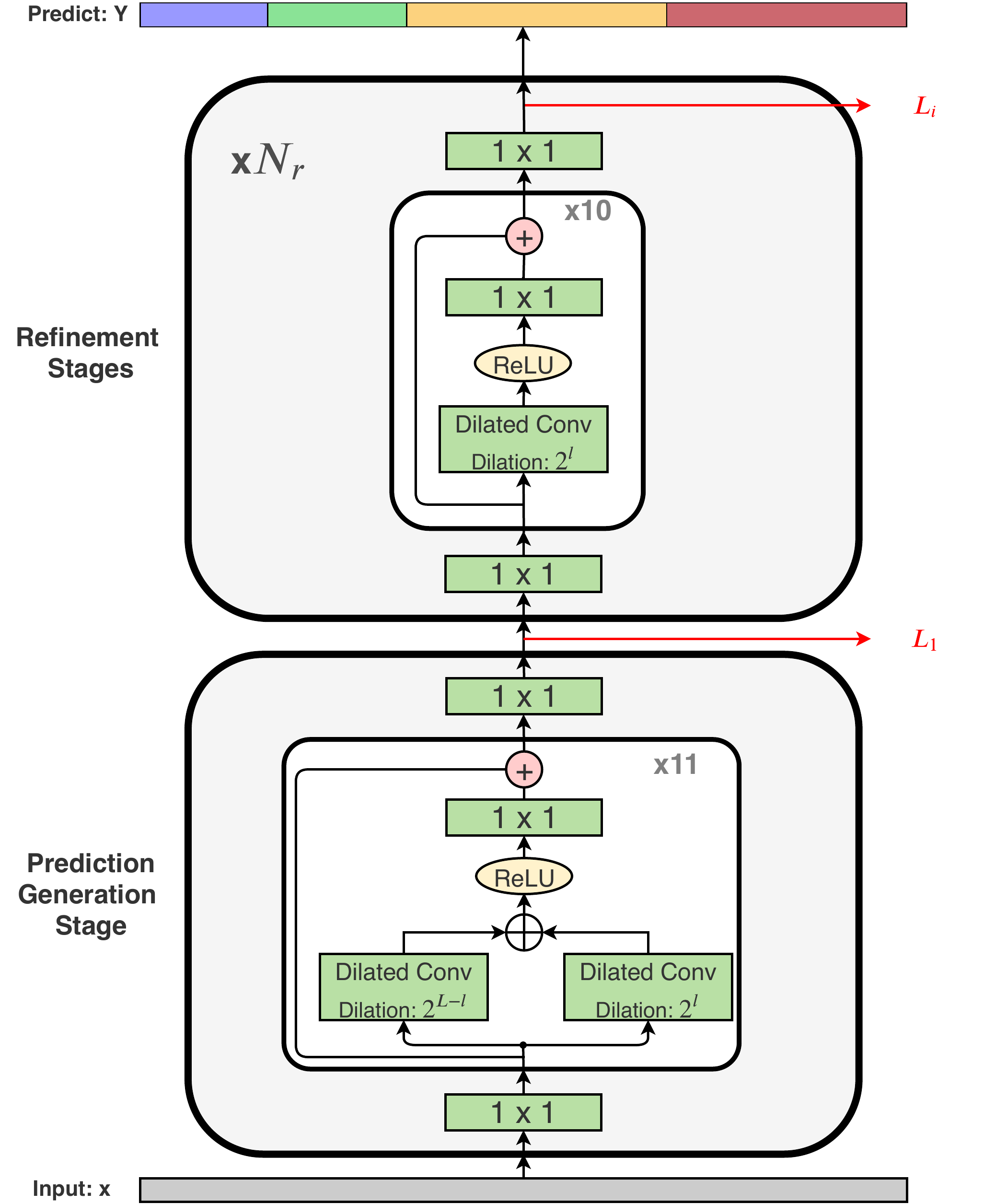}
   \caption{Overview of MS-TCN++. The first stage adapts an SS-TCN model with dual dilated 
   layers. This stage generates an initial prediction that is refined incrementally by a set 
   of $N_r$ refinement stages. For the refinement stages, an SS-TCN with dilated residual layers is 
   used. A loss layer is added after each stage.}
\label{fig:ms-tcn++}
\end{figure}

In this section, we introduce MS-TCN++, which utilizes the proposed dual dilated layer to improve 
MS-TCN. Similar to MS-TCN, the first stage in MS-TCN++ is responsible for generating the initial prediction, 
whereas the remaining stages incrementally refine this prediction.
For the prediction generation stage, we adapt an SS-TCN with dual dilated layers (Figure~\ref{fig:ddl}) 
replacing the simple dilated residual layers (Figure~\ref{fig:layer}) that are originally used in SS-TCN. 
Using the DDL enables the prediction generation stage to capture both local and global features in all layers, 
which results in better predictions. As refinement is easier than prediction generation, we adapt the SS-TCN 
architecture with dilated residual layers for the refinement stages. In our experiments, we show that using 
DDL only for the first stage performs best. Figure~\ref{fig:ms-tcn++} shows an overview of the proposed MS-TCN++.

While adding more stages incrementally refines the predictions, it also drastically increases 
the number of parameters. Nevertheless, as the refinement stages are sharing the same role, their 
parameters can be shared to get a more compact model. In the experiments, we show that sharing the 
parameters between the refinement stages significantly reduces the number of parameters with only 
a slight degradation in accuracy.

\subsection{Loss Function}
\label{sec:loss_function}

As a loss function, we use a combination of a classification loss and a smoothing 
loss. For the classification loss, we use a cross entropy loss
\begin{equation}
\mathcal{L}_{cls} = \frac{1}{T}\sum_{t} -log(y_{t,c}), 
\end{equation}
where $y_{t,c}$ is the predicted probability for the ground truth 
label $c$ at time $t$.

While the cross entropy loss already performs well, we found that the predictions 
for some of the videos contain a few over-segmentation errors. To further improve 
the quality of the predictions, we use an additional smoothing loss to reduce such 
over-segmentation errors. For this loss, we use a truncated mean squared error over 
the frame-wise log-probabilities

\begin{equation}\label{eqn:t_mse}
\mathcal{L}_{T-MSE} = \frac{1}{TC}\sum_{t,c}\tilde{\Delta}_{t,c}^2, 
\end{equation}
\begin{equation}
\tilde{\Delta}_{t,c} = 
\begin{cases}
\Delta_{t,c}        &: \Delta_{t,c} \leq \tau\\
\tau                &: otherwise\\
\end{cases}, 
\end{equation}
\begin{equation}
\Delta_{t,c} = \left| log\ y_{t,c} - log\ y_{t-1,c} \right|,
\end{equation}
where $T$ is the video length, $C$ is the number of classes, and $y_{t,c}$ 
is the probability of class $c$ at time $t$. 

Note that the gradients are only computed 
with respect to $y_{t,c}$, whereas $y_{t-1,c}$ is not considered as a function 
of the model's parameters. This loss is similar to the Kullback-Leibler (KL) 
divergence loss where 
\begin{equation}
\mathcal{L}_{KL} = \frac{1}{T}\sum_{t,c} y_{t-1,c} (log\ y_{t-1,c} - log\ y_{t,c}).
\end{equation}
However, we found that the truncated mean squared error ($\mathcal{L}_{T-MSE}$)~\eqref{eqn:t_mse} 
reduces the over-segmentation errors more. We will compare the KL loss and the proposed 
loss in the experiments.

The final loss function for a single stage is a combination of the 
above mentioned losses
\begin{equation}
\mathcal{L}_s = \mathcal{L}_{cls} + \lambda \mathcal{L}_{T-MSE}, 
\end{equation}
where $\lambda$ is a model hyper-parameter to determine the contribution of 
the different losses. Finally to train the complete model, we minimize the 
sum of the losses over all stages
\begin{equation}
\mathcal{L} = \sum_s \mathcal{L}_{s} . 
\end{equation}

\subsection{Implementation Details}
\label{sec:implementation details}
For both MS-TCN and MS-TCN++, we use a multi-stage architecture with four stages. While all stages are the same for 
MS-TCN, the stages in MS-TCN++ consist of one prediction generation stage and three refinement stages. 
Each stage in MS-TCN and the refinement stages in MS-TCN++ contain ten dilated convolution layers. For the prediction 
generation stage in MS-TCN++, we use eleven layers. 
Dropout is used after each layer with probability $0.5$. We set the number of filters to $64$ in all layers 
of the model and the filter size is $3$. For the loss function, we set $\tau = 4$ and $\lambda = 0.15$. 
In all experiments, we use Adam optimizer with a learning rate of $0.0005$.

\section{Experiments}
\noindent\textbf{Datasets.} We evaluate the proposed models on three challenging datasets: 50Salads~\cite{stein2013combining}, Georgia Tech Egocentric Activities (GTEA)~\cite{fathi2011learning}, and the Breakfast dataset~\cite{kuehne2014language}. Table~\ref{tab:dataset} 
shows a summary of these datasets.

\begin{table*}[tb]
\centering
\resizebox{\linewidth}{!}{%
\begin{tabular}{lcccccc}
\toprule
          & \# videos  & \# action classes  & view & description  \\
\midrule 
50Salads  &      50   &         17        & top-view & salad preparation 
activities \\
GTEA      &      28   &         11         & egocentric & 7 different activities, like preparing coffee or cheese sandwich \\
Breakfast &     1712  &         48        & third person view & breakfast preparation related activities in 18 different kitchens \\
\bottomrule

\end{tabular}%
}
\vspace{1mm}
\caption{Summary of the used datasets in the experiments.}
\label{tab:dataset}
\end{table*}

The \textbf{50Salads} dataset contains 50 videos with $17$ action classes. 
The videos were recorded from the top view. 
On average, each video contains 20 action instances and is $6.4$ minutes long. 
As the name of the dataset indicates, the videos depict salad preparation 
activities. 
These activities were performed by $25$ actors where each actor 
prepared two different salads. 
For evaluation, we use five-fold cross-validation and report the average as in~\cite{stein2013combining}. 

The \textbf{GTEA} dataset contains $28$ videos corresponding to 7 different activities, like preparing coffee or cheese sandwich, performed by 4 subjects. 
This dataset contains egocentric videos recorded by a camera that is mounted on the actor's head. 
The frames of the videos are annotated with $11$ action classes including 
background. 
On average, each video has 20 action instances. 
We use cross-validation for evaluation by leaving one subject out. 

The \textbf{Breakfast} dataset is the largest among the three datasets with $1,712$ third person view videos. 
The videos were recorded in 18 different kitchens showing breakfast preparation related activities. 
Overall, there are $48$ different actions where each video contains $6$ action instances on average. 
For evaluation, we use the standard 4 splits as proposed in~\cite{kuehne2014language} and report the average.

For all datasets, we extract I3D~\cite{carreira2017quo} features for the video frames and use these features as input to our model. 
For the GTEA and Breakfast datasets we use the temporal video resolution at $15$ fps, while for 50Salads we downsampled the features from $30$ fps to $15$ fps to be consistent with the other datasets.
\\

\noindent\textbf{Evaluation Metrics.} For evaluation, we report the frame-wise accuracy (Acc), segmental edit distance and the segmental F1 score at overlapping thresholds $10\%,\ 25\%$ and $50\%$, denoted by $F1$@$\{10,25,50\}$. 
The overlapping threshold is determined based on the intersection over union (IoU) ratio. 
While the frame-wise accuracy is the most commonly used metric for action segmentation, long action classes have a higher impact than short action classes on this metric and over-segmentation errors have a very low impact.
For that reason, we use the segmental F1 score as a measure of the quality of the prediction as proposed by~\cite{lea2017temporal}.

\subsection{Effect of the Number of Stages}

\begin{figure}[tb]
   \includegraphics[trim={1cm 1.5cm 1cm 2.5cm},clip,width=.95\linewidth]{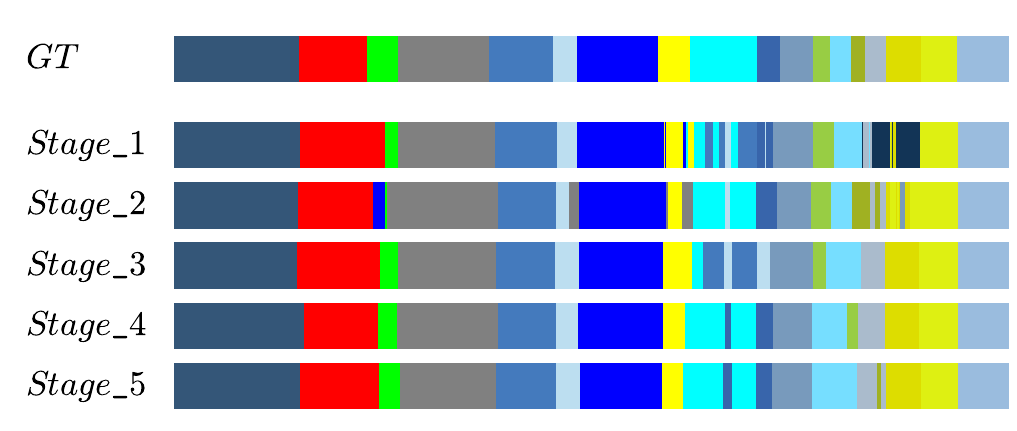}
   \caption{Qualitative result from the 50Salads dataset for comparing different number of stages.}
\label{fig:qualitative_results_stages}
\end{figure}

We start our evaluation by showing the effect of using a multi-stage architecture (MS-TCN). 
Table~\ref{tab:number_of_stages} shows the results of a single-stage model compared to multi-stage models with different number of stages. 
As shown in the table, all of these models achieve a comparable frame-wise accuracy. 
Nevertheless, the quality of the predictions is very different. 
Looking at the segmental edit distance and F1 scores of these models, we can see that the single-stage model produces a lot of over-segmentation errors, as indicated by the low F1 score. 
On the other hand, using a multi-stage architecture reduces these errors and increases the F1 score. 
This effect is clearly visible when we use two or three stages, which gives a huge boost to the accuracy. 
Adding the fourth stage still improves the results but not as significant as the previous stages. 
However, by adding the fifth stage, we can see that the performance starts 
to degrade.
This might be an over-fitting problem as a result of increasing the number of parameters. 
The effect of the multi-stage architecture can also be seen in the qualitative results shown in Figure~\ref{fig:qualitative_results_stages}. 
Adding more stages results in an incremental refinement of the predictions.
In the rest of the experiments we use a multi-stage TCN with four stages.

\begin{table}[tb]
\centering
\resizebox{.95\linewidth}{!}{%
\begin{tabular}{lccccc}
\toprule
 & \multicolumn{3}{c}{F1@\{10,25,50\}} & Edit & Acc  
\\ \midrule
SS-TCN            &         27.0  &         25.3  &         21.5  &         20.5  &         78.2  \\
MS-TCN (2 stages) &         55.5  &         52.9  &         47.3  &         47.9  &         79.8  \\
MS-TCN (3 stages) &         71.5  &         68.6  &         61.1  &         64.0  &         78.6  \\ 
MS-TCN (4 stages) &         76.3  & \textbf{74.0} & \textbf{64.5} &         67.9  & \textbf{80.7} \\ 
MS-TCN (5 stages) & \textbf{76.4} &         73.4  &         63.6  & \textbf{69.2} &         79.5  \\   
\bottomrule

\end{tabular}%
}
\vspace{1mm}
\caption{Effect of the number of stages on the 50Salads dataset.}
\label{tab:number_of_stages}
\end{table}

\subsection{Multi-Stage TCN vs. Deeper Single-Stage TCN}

In the previous section, we have seen that our multi-stage architecture is 
better than a single-stage one. 
However, that comparison does not show whether  the improvement is because of the multi-stage architecture or due to the increase in the number of parameters when adding more stages. 
For a fair comparison, we train a single-stage model that has the same number of parameters as the multi-stage one. 
As each stage in our MS-TCN contains $12$ layers (ten dilated convolutional layers, one $1 \times 1$ convolutional layer and a softmax layer), we train a single-stage TCN with $48$ layers, which is the number of layers in a MS-TCN with four stages. 
For the dilated convolutions, we use similar dilation factors as in our MS-TCN. \Ie. we start with a dilation factor of $1$ and double it at every layer up to a factor of $512$, and then we start again from $1$. 
As shown in Table~\ref{tab:more_layers}, our multi-stage architecture outperforms its single-stage counterpart with a large margin of up to $27\%$. This highlights the impact of the multi-stage architecture in improving the quality of the predictions.

\begin{table}[tb]
\centering
\resizebox{.95\linewidth}{!}{%
\begin{tabular}{lccccc}
\toprule
  & \multicolumn{3}{c}{F1@\{10,25,50\}} & Edit & Acc  
\\ \midrule
SS-TCN (48 layers) &         49.0  &         46.4  &         40.2  &         40.7  &         78.0  \\
MS-TCN             & \textbf{76.3} & \textbf{74.0} & \textbf{64.5} & \textbf{67.9} & \textbf{80.7} \\ 
\bottomrule
\end{tabular}%
}
\vspace{1mm}
\caption{Comparing a multi-stage TCN with a deep single-stage TCN on the 50Salads dataset.}
\label{tab:more_layers}
\end{table}

\subsection{Comparing Different Loss Functions}

\begin{table}[tb]
\centering
\resizebox{.9\linewidth}{!}{%
\begin{tabular}{lccccc}
\toprule
  & \multicolumn{3}{c}{F1@\{10,25,50\}} & Edit & Acc  
\\ \midrule
$\mathcal{L}_{cls} $       							 &        71.3 &        69.7 &        60.7 &        64.2 &        79.9  \\
$\mathcal{L}_{cls} + \lambda \mathcal{L}_{KL}$	     &        71.9 &        69.3 &        60.1 &        64.6 &        80.2  \\
$\mathcal{L}_{cls} + \lambda \mathcal{L}_{T-MSE}$    &\textbf{76.3}&\textbf{74.0}&\textbf{64.5}&\textbf{67.9}&\textbf{80.7} \\ 
\bottomrule
\end{tabular}%
}
\vspace{1mm}
\caption{Comparing different loss functions on the 50Salads dataset.}
\label{tab:loss_function}
\end{table}

\begin{figure}[tb]
   \includegraphics[trim={.4cm 2.7cm 1cm 2.5cm},clip,width=.95\linewidth]{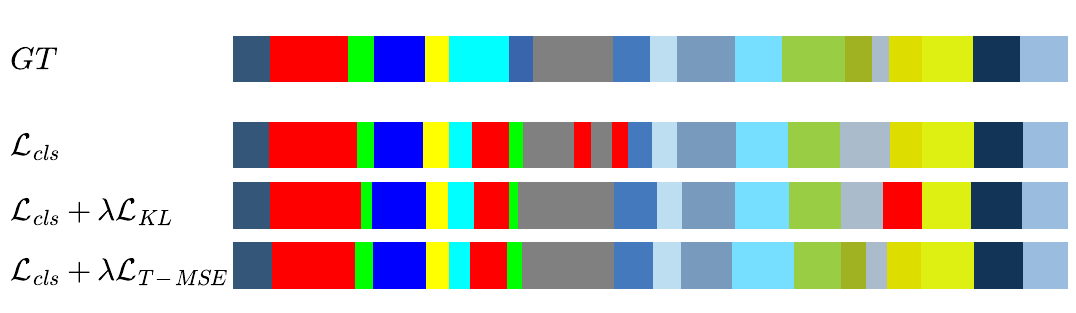}
   \caption{Qualitative result from the 50Salads dataset for comparing different loss functions.}
\label{fig:loss_functions}
\end{figure}

As a loss function, we use a combination of a cross-entropy loss, which 
is common practice for classification tasks, and a truncated mean squared 
loss over the frame-wise log-probabilities to ensure smooth predictions. 
While the smoothing loss slightly improves the frame-wise accuracy compared to 
the cross entropy loss alone, we found that this loss produces much less 
over-segmentation errors. 
Table~\ref{tab:loss_function} and Figure~\ref{fig:loss_functions} show a comparison of these losses. 
As shown in Table~\ref{tab:loss_function}, the proposed loss achieves 
better F1 and edit scores with an absolute improvement of up to $5\%$. 
This indicates that our loss produces less over-segmentation errors compared 
to cross entropy since it forces consecutive frames to have similar class probabilities, which results in a smoother output. 

\begin{figure}[tb]
\begin{center}
\begin{tabular}{cc}
   \includegraphics[trim={4cm .6cm 27cm .3cm},clip,width=.44\linewidth]{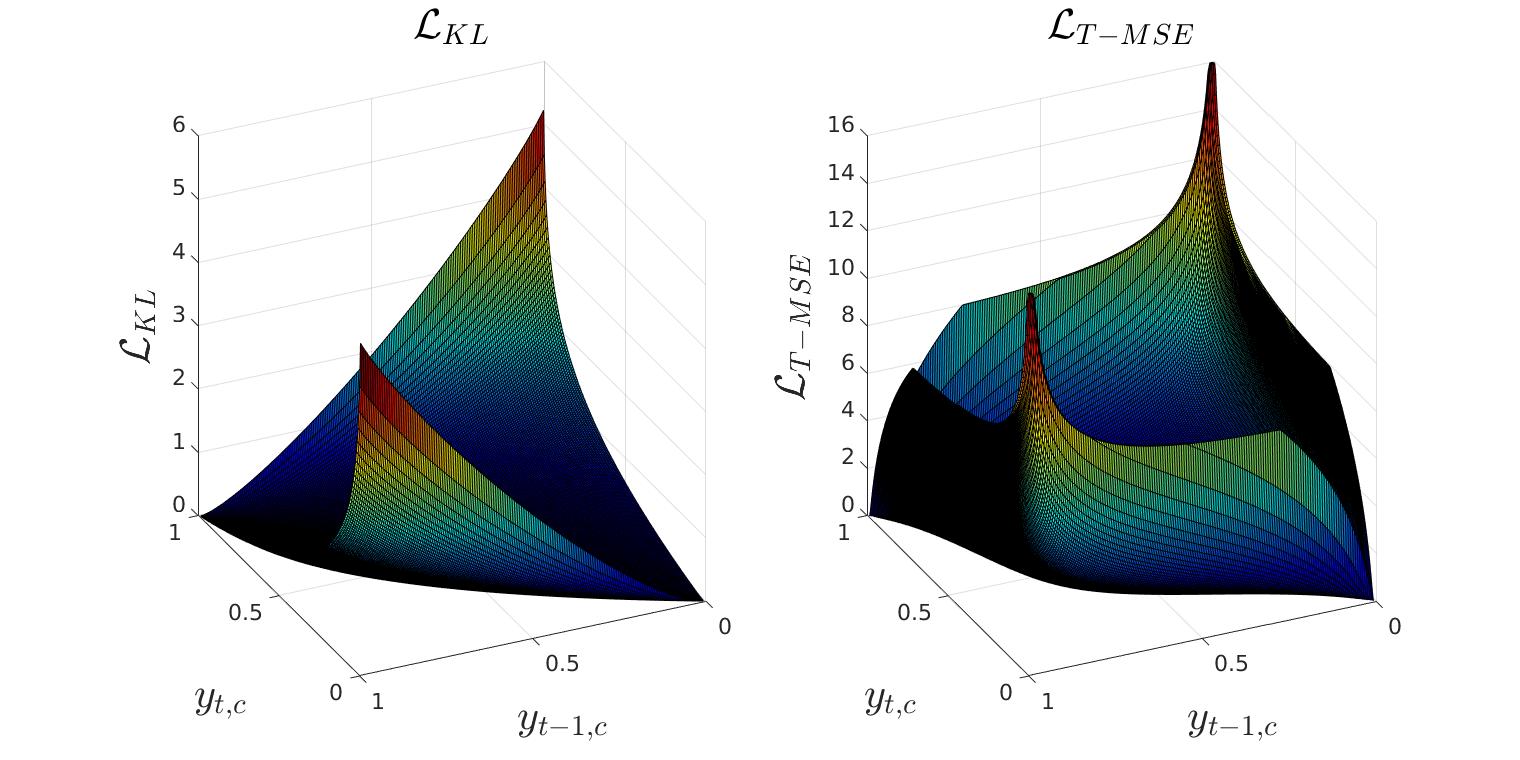} & 
   \includegraphics[trim={27cm .6cm 4cm .3cm},clip,width=.44\linewidth]{losses_plot.jpg}
\end{tabular}
\end{center}
   \caption{Loss surface for the Kullback-Leibler (KL) divergence loss ($\mathcal{L}_{KL}$) 
   and the proposed truncated mean squared loss ($\mathcal{L}_{T-MSE}$) for the case of two 
   classes. $y_{t,c}$ is the predicted probability for class c and $y_{t-1,c}$ is the target 
   probability corresponding to that class.}
\label{fig:losses_plot}
\vspace{3mm}
\end{figure}

Penalizing the difference in log-probabilities is similar to the Kullback-Leibler (KL) divergence loss, which measures the difference between two probability distributions. 
However, the results show that the proposed loss produces better results than the KL loss as shown in Table~\ref{tab:loss_function} and Figure~\ref{fig:loss_functions}. 
The reason behind this is the fact that the KL divergence loss does not penalize cases where the difference between the target probability and the predicted probability is very small. 
Whereas the proposed loss penalizes small differences as well. 
Note that, in contrast to the KL loss, the proposed loss is symmetric. Figure~\ref{fig:losses_plot} shows the surface for both the KL loss and the proposed truncated mean squared loss for the case of two classes. 
We also tried a symmetric version of the KL loss but it performed worse than the KL loss.

\subsection{Impact of $\lambda$ and $\tau$}

The effect of the proposed smoothing loss is controlled by two hyper-parameters: 
$\lambda$ and $\tau$. In this section, we study the impact of these parameters and see how they affect the performance of the proposed model.

\noindent\textbf{Impact of $\lambda$:} In all experiments, we set $\lambda=0.15$. 
To analyze the effect of this parameter, we train different models with different values of $\lambda$. 
As shown in Table~\ref{tab:impact_lambda_tau}, the impact of $\lambda$ is very small on the performance. 
Reducing $\lambda$ to $0.05$ still improves the performance but not as good as the default value of $\lambda=0.15$. 
Increasing its value to $\lambda=0.25$ also causes a degradation in 
performance. 
This drop in performance is due to the fact that the smoothing loss heavily penalizes changes in frame-wise labels, which affects the detected boundaries 
between action segments.

\noindent\textbf{Impact of $\tau$:} This hyper-parameter defines the threshold to truncate 
the smoothing loss. Our default value is $\tau=4$. While reducing the value to $\tau=3$ 
still gives an improvement over the cross entropy baseline, setting $\tau=5$ results 
in a huge drop in performance. This is mainly because when $\tau$ is too high, the smoothing 
loss penalizes cases where the model is very confident that the consecutive frames 
belong to two different classes, which indeed reduces the capability of the model in 
detecting the true boundaries between action segments.

\begin{table}[tb]
\centering
\resizebox{\linewidth}{!}{%
\begin{tabular}{lccccc}
\toprule
\textbf{Impact of $\lambda$}  & \multicolumn{3}{c}{F1@\{10,25,50\}} & Edit & Acc  
\\ \midrule
MS-TCN ($\lambda=0.05,\ \tau=4$)   &        74.1 &        71.7 &        62.4 &        66.6 &        80.0  \\ 
MS-TCN ($\lambda=0.15,\ \tau=4$)   &\textbf{76.3}&\textbf{74.0}&\textbf{64.5}&        67.9 &\textbf{80.7} \\ 
MS-TCN ($\lambda=0.25,\ \tau=4$)   &        74.7 &        72.4 &        63.7 &\textbf{68.1}&        78.9  \\ 
\bottomrule
\toprule
\textbf{Impact of $\tau$}  & \multicolumn{3}{c}{F1@\{10,25,50\}} & Edit & Acc  
\\ \midrule
MS-TCN ($\lambda=0.15,\ \tau=3$)   &        74.2 &        72.1 &        62.2 &        67.1 &        79.4  \\ 
MS-TCN ($\lambda=0.15,\ \tau=4$)   &\textbf{76.3}&\textbf{74.0}&\textbf{64.5}&\textbf{67.9}&\textbf{80.7} \\ 
MS-TCN ($\lambda=0.15,\ \tau=5$)   &        66.6 &        63.7 &        54.7 &        60.0 &        74.0  \\ 
\bottomrule
\end{tabular}%
}
\vspace{1mm}
\caption{Impact of $\lambda$ and $\tau$ on the 50Salads dataset.}
\label{tab:impact_lambda_tau}
\end{table}

\subsection{Effect of Passing Features to Higher Stages}

\begin{table}[tb]
\centering
\resizebox{\linewidth}{!}{%
\begin{tabular}{lccccc}
\toprule
  & \multicolumn{3}{c}{F1@\{10,25,50\}} & Edit & Acc  
\\ \midrule
Probabilities and features 	&        56.2 &        53.7 &        45.8 &        47.6 &       76.8  \\
Probabilities only          &\textbf{76.3}&\textbf{74.0}&\textbf{64.5}&\textbf{67.9}&\textbf{80.7} \\
\bottomrule
\end{tabular}%
}
\vspace{1mm}
\caption{Effect of passing features to higher stages on the 50Salads dataset.}
\label{tab:features}
\end{table}



\begin{figure}[tb]
\begin{center}
   \includegraphics[trim={0cm 5.4cm 1cm 4.05cm},clip,width=\linewidth]{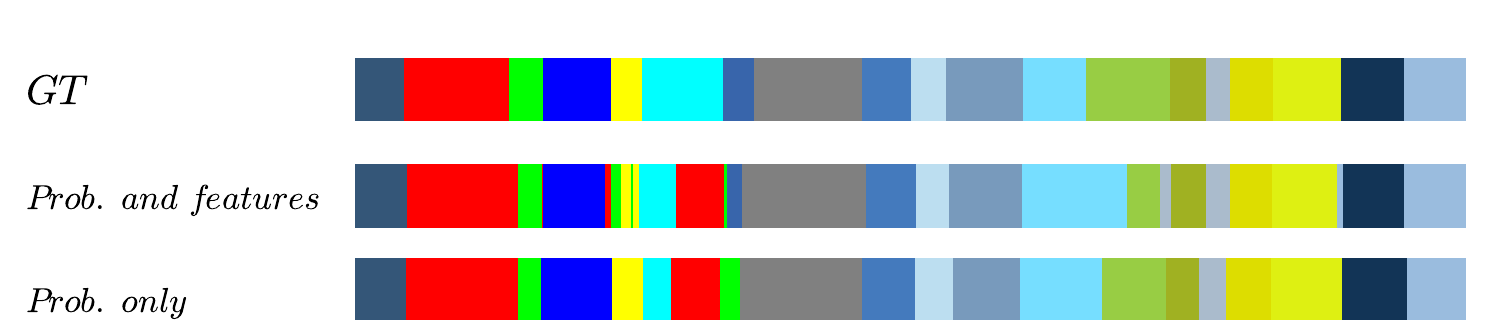}
   \\
   \includegraphics[trim={0cm 5.4cm 1cm 4.05cm},clip,width=\linewidth]{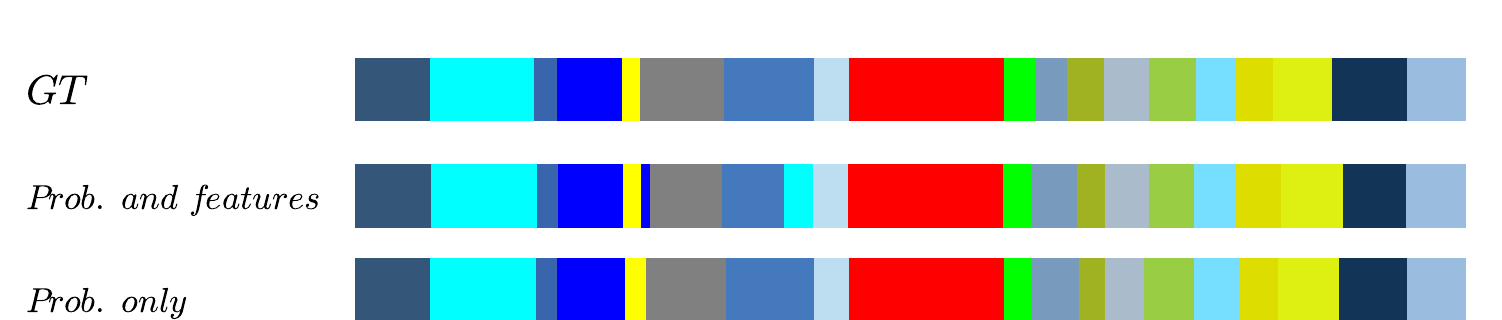}
\end{center}
   \caption{Qualitative results for two videos from the 50Salads dataset showing the effect 
   of passing features to higher stages.}
\label{fig:features}
\end{figure}

In the proposed multi-stage TCN, the input to higher stages are the 
frame-wise probabilities only. However, in the multi-stage architectures 
that are used for human pose estimation, additional features are 
usually concatenated to the output heat-maps of the previous stage.  
In this experiment, we therefore analyze the effect of combining additional 
features to the input probabilities of higher stages. To this end, we trained 
two multi-stage TCNs: one only with the predicted frame-wise probabilities 
as input to the next stage, and, for the second model, we concatenated the output 
of the last dilated convolutional layer in each stage to the input probabilities 
of the next stage. As shown in Table~\ref{tab:features}, concatenating 
the features to the input probabilities results in a huge drop of the F1 score 
and the segmental edit distance (around $20\%$). We argue that the reason behind this 
degradation in performance is that a lot of action classes share similar appearance 
and motion. By adding the features of such classes at each stage, the model 
is confused and produces small separated falsely detected action segments 
that correspond to an over-segmentation effect. Passing only the probabilities 
forces the model to focus on the context of neighboring labels, which are explicitly 
represented by the probabilities. This effect can also be seen in the qualitative 
results shown in Figure~\ref{fig:features}.


\subsection{MS-TCN++ vs. MS-TCN}
\begin{figure}[tb]
\begin{center}
   \includegraphics[trim={0cm 5.4cm 1cm 4.05cm},clip,width=\linewidth]{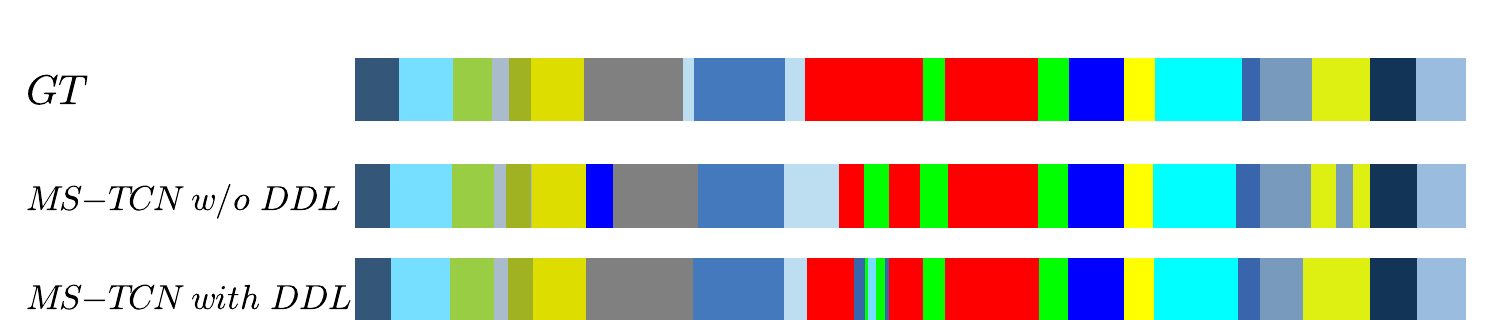}
   \\
   \includegraphics[trim={0cm 5.4cm 1cm 4.05cm},clip,width=\linewidth]{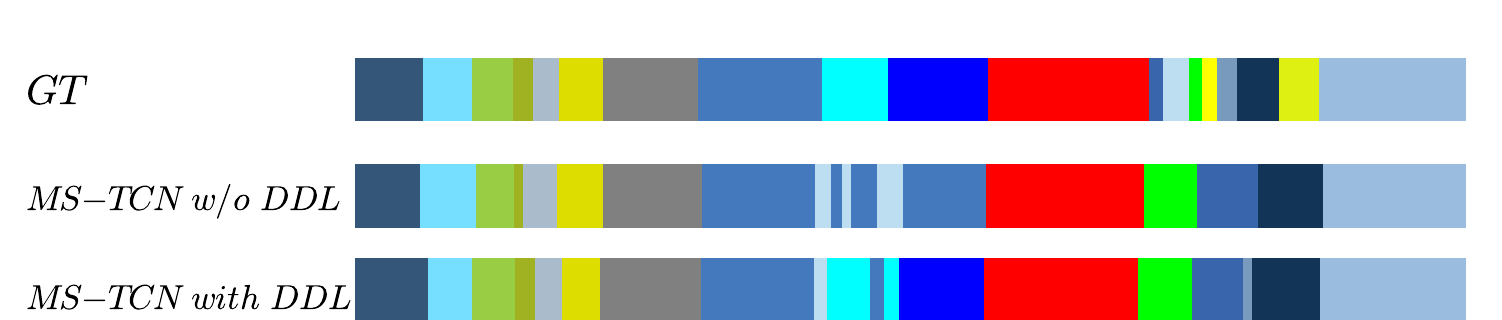}
\end{center}
   \caption{Qualitative results for two videos from the 50Salads dataset showing the impact 
   of the dual dilated layer (DDL).}
\label{fig:ddl_q_res}
\end{figure}

In this section, we compare the two multi-stage architectures: MS-TCN++ and MS-TCN. In contrast to 
MS-TCN, MS-TCN++ uses the dual dilated layer (DDL) in the first stage. Table~\ref{tab:tcn_vs_tcnpp} 
shows the results of both architectures on the 50Salads dataset. As shown in the table, MS-TCN++ 
outperforms MS-TCN with a large margin of up to $6.4\%$. This emphasizes the importance of combining both 
local and global representations in the prediction generation stage by utilizing the DDL in MS-TCN++. 
To study the impact of using DDL in all stages, we also train an MS-TCN where we use the DDL in all stages. 
As shown in Table~\ref{tab:tcn_vs_tcnpp}, MS-TCN++ outperforms MS-TCN with DDL in all stages. This 
indicates that decoupling the design of the refinement stages and the prediction generation stage is 
crucial. While utilizing the global context by the DDL is crucial for the prediction generation stage, the refinement stages focus more on the local context. By adding DDL to the refinement stages, the accuracy even drops due to overfitting. Note that using DDL in all stages outperforms MS-TCN with up to $2.8\%$. This further highlights 
the gains of the DDL. The impact of DDL is also visible in the qualitative results shown in Figure~\ref{fig:ddl_q_res}.

\begin{table}[tb]
\centering
\resizebox{\linewidth}{!}{%
\begin{tabular}{lccccc}
\toprule
  & \multicolumn{3}{c}{F1@\{10,25,50\}} & Edit & Acc  
\\ \midrule
MS-TCN      	&        76.3 &        74.0 &    64.5     &       67.9  &       80.7  \\
MS-TCN with DDL &        77.3 &        75.0 &    67.3     &       69.8  &       82.4  \\
MS-TCN++        &\textbf{80.7}&\textbf{78.5}&\textbf{70.1} & \textbf{74.3} & \textbf{83.7}   \\
\bottomrule
\end{tabular}%
}
\caption{MS-TCN++ vs. MS-TCN vs. MS-TCN with DDL on the 50Salads dataset.}
\label{tab:tcn_vs_tcnpp}
\end{table}


\subsection{Impact of the Number of Layers}
For MS-TCN and the refinement stages in MS-TCN++, we fix the number of layers in each stage to $10$ layers. 
Whereas for the prediction generation stage in MS-TCN++, we set the number of layers to $11$. In this section, 
we study the impact of these parameters. Table~\ref{tab:number_of_layers} shows the impact of the number of 
layers ($L$) for the MS-TCN stages on the 50Salads dataset. Increasing $L$ form $8$ to $10$ significantly 
improves the performance. This is mainly due to the increase in the receptive field. Using more than $10$ 
layers ($L = 11$, $L = 12$) does not improve the frame-wise accuracy but slightly increases the F1 scores. 
We also tried to change the number of layers only in the refinement stages in MS-TCN. As shown in 
Table~\ref{tab:number_of_layers_rnet}, this does not have a significant impact and using $10$ layers achieves 
the best performance.
Also for the refinement stages in MS-TCN++, the number of layers $L_r$ does not have a significant impact on the 
performance. To be consistent with~\cite{abufarha2019tcn}, we set $L_r = 10$ which achieves a reasonable 
trade-off on performance with respect to all evaluation metrics as shown in Table~\ref{tab:number_of_layers_rnet}. 
A similar behavior can be observed in Table~\ref{tab:number_of_layers_bg_net} 
for the number of layers $L_g$ in the prediction generation stage with $L_g = 11$ achieving 
the best performance. Generally speaking, the number of layers in each stage has more impact in MS-TCN 
compared to MS-TCN++. As the main difference between these two models is the dual dilation layer (DDL) 
that is used in MS-TCN++, this indicates that the DDL can better capture both local and global features 
to generate much better predictions.

\begin{table}[tb]
\centering
\resizebox{.87\linewidth}{!}{%
\begin{tabular}{lccccc}
\toprule
  & \multicolumn{3}{c}{F1@\{10,25,50\}} & Edit & Acc  
\\ \midrule
$L$ = 6      &         53.2  &         48.3  &         39.0  &         46.2  &         63.7  \\
$L$ = 8      &         66.4  &         63.7  &         52.8  &         60.1  &         73.9  \\
$L$ = 10     &         76.3  &         74.0  &         64.5  &         67.9  & \textbf{80.7}  \\
$L$ = 11     &         76.7 &         74.2 &         65.5 &   \textbf{69.7} &         80.4   \\
$L$ = 12     & \textbf{77.8} & \textbf{75.2} & \textbf{66.9} &         69.6 &         80.5 \\ 
\bottomrule
\end{tabular}%
}
\caption{Effect of the number of layers ($L$) in each stage of MS-TCN on the 50Salads dataset.}
\label{tab:number_of_layers}
\end{table}

\begin{table}[tb]
\centering
\resizebox{.95\linewidth}{!}{%
\begin{tabular}{llccccc}
\toprule
  & \multicolumn{3}{c}{F1@\{10,25,50\}} & Edit & Acc  
\\ \midrule
         & $L_r$ = 6      &         74.3  &         71.5  &         62.8  &         66.0  &         78.6  \\
         & $L_r$ = 8      &         75.4  &         72.4  &         64.3  & \textbf{68.0} &         79.5  \\
MS-TCN   & $L_r$ = 10     & \textbf{76.3} & \textbf{74.0} & \textbf{64.5} &         67.9  & \textbf{80.7}  \\
         & $L_r$ = 11     &         75.0  &         72.0  &         63.5  &         67.6  &         80.3  \\
         & $L_r$ = 12     &         74.1  &         71.2  &         62.3  &         65.7  &         79.1  \\
\midrule
         & $L_r$ = 6      &         78.2  &         75.6  &         67.7  &         69.6  &         82.3  \\
         & $L_r$ = 8      & \textbf{80.9} &         78.2  & \textbf{70.2} &         73.4  &        82.9  \\
MS-TCN++ & $L_r$ = 10     &         80.7 & \textbf{78.5} &          70.1  &  \textbf{74.3} & \textbf{83.7}   \\
         & $L_r$ = 11     &         80.5 &         78.3  &          70.0  &          72.6 &          83.4   \\
         & $L_r$ = 12     &         79.4 &         76.9 &           69.2  &          71.3  &         83.5 \\ 
\bottomrule
\end{tabular}%
}
\caption{Effect of the number of layers ($L_r$) in each refinement stage on the 50Salads dataset.}
\label{tab:number_of_layers_rnet}
\end{table}

\begin{table}[tb]
\centering
\resizebox{.87\linewidth}{!}{%
\begin{tabular}{lccccc}
\toprule
  & \multicolumn{3}{c}{F1@\{10,25,50\}} & Edit & Acc  
\\ \midrule
$L_g$ = 6      &         74.3  &         71.6  &         63.5  &         67.8  &         78.5  \\
$L_g$ = 8      &         77.4  &         75.3  &         67.8  &         70.3  &         80.8  \\
$L_g$ = 10     &         79.8  &         77.9  &\textbf{71.0}  &         72.5  &         83.1  \\
$L_g$ = 11     & \textbf{80.7} & \textbf{78.5} &          70.1 &  \textbf{74.3} & \textbf{83.7}   \\
$L_g$ = 12     &         78.9 &          76.6 &          67.6 &          70.8  &         83.2 \\ 
\bottomrule
\end{tabular}%
}
\caption{Effect of the number of layers ($L_g$) in the prediction generation stage for MS-TCN++ on the 50Salads dataset.}
\label{tab:number_of_layers_bg_net}
\end{table}

\subsection{Impact of the Large Receptive Field on Short Videos}

To study the impact of the large receptive field on short videos, we 
evaluate MS-TCN and MS-TCN++ on three groups of videos based on their durations. 
For this evaluation, we use the GTEA dataset since it contains shorter videos 
compared to the other datasets. As shown in Table~\ref{tab:impact_on_short_videos}, 
both MS-TCN and MS-TCN++ perform well on both short and long videos. Nevertheless, the performance is 
slightly worse on longer videos due to the limited receptive field. The improvements of MS-TCN++ over 
MS-TCN are also noticeable for both short and long videos.

\begin{table}[tb]
\centering
\resizebox{.95\linewidth}{!}{%
\begin{tabular}{llcccccc}
\toprule
& Duration & \multicolumn{3}{c}{F1@\{10,25,50\}} & Edit & Acc  
\\ \midrule
       & $< 1$ min        &         89.6  &         87.9  &         77.0  &         82.5  &         76.6  \\
MS-TCN & $1 - 1.5$ min    &         85.9  &         84.3  &         71.9  &         80.7  &         76.4  \\
       & $\geq 1.5$ min   &         81.2  &         76.5  &         58.4  &         71.8  &         75.9 

\\ \midrule
         & $< 1$ min        &         90.4  &         90.4  &         80.8  &         84.4  &         79.3  \\
MS-TCN++ & $1 - 1.5$ min    &         88.7  &         85.8  &         75.1  &         83.6  &         79.3  \\
         & $\geq 1.5$ min   &         80.8  &         78.8  &         63.3  &         76.1  &         77.2  
\\ 
\bottomrule

\end{tabular}%
}
\vspace{1mm}
\caption{Evaluation of three groups of videos from the GTEA dataset based on their durations.}
\label{tab:impact_on_short_videos}
\end{table}


\begin{figure*}[tb]
\centering
\subfigure[]{
\includegraphics[trim={.5cm 4cm .5cm .5cm},clip,width=.47\linewidth]{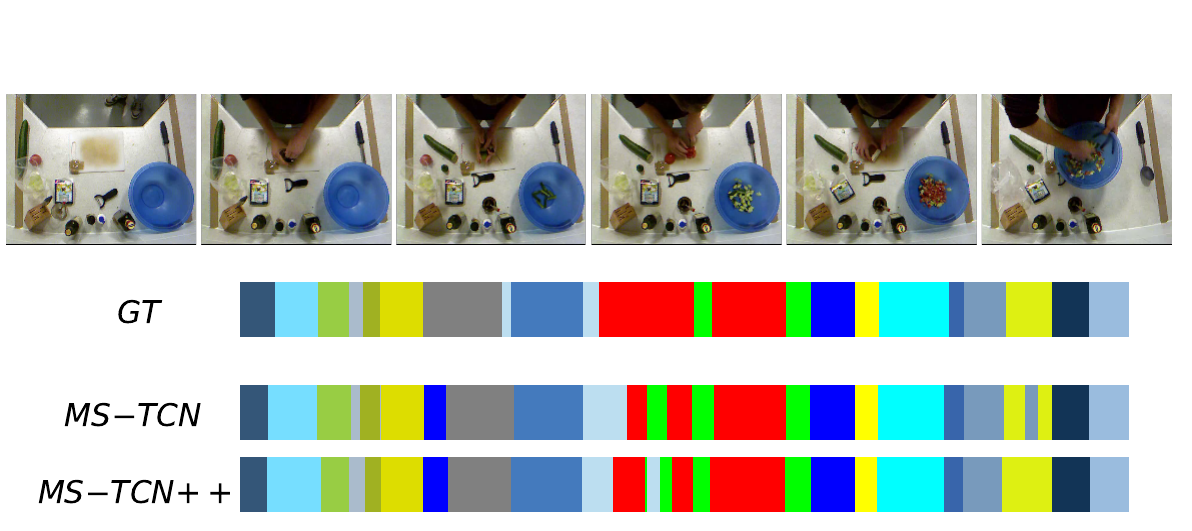}
}
\quad
\subfigure[]{
\includegraphics[trim={.5cm 4cm .5cm .5cm},clip,width=.47\linewidth]{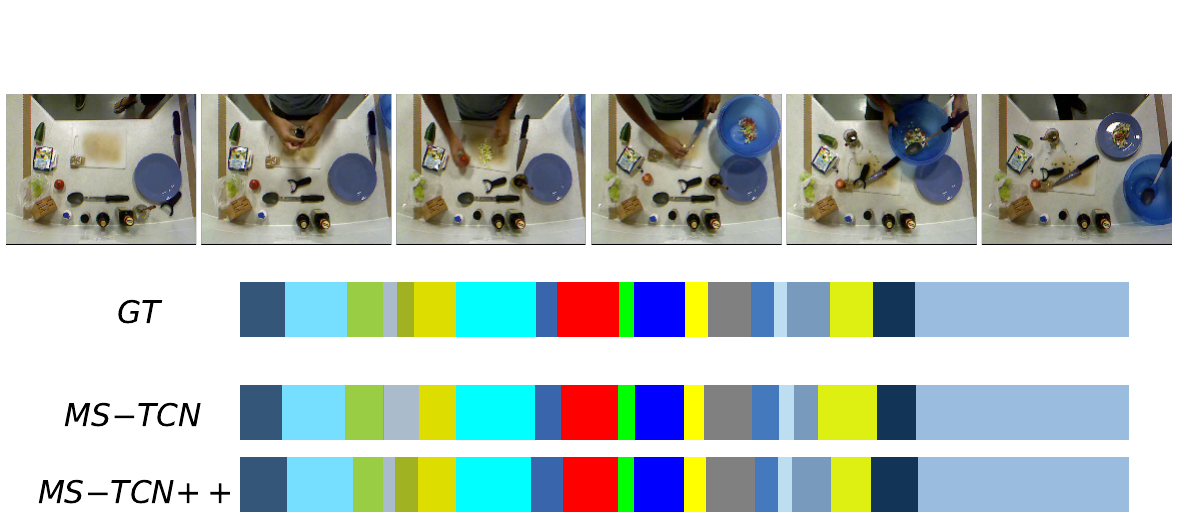}
}
\quad
\subfigure[]{
\includegraphics[trim={.5cm 4cm .5cm .5cm},clip,width=.47\linewidth]{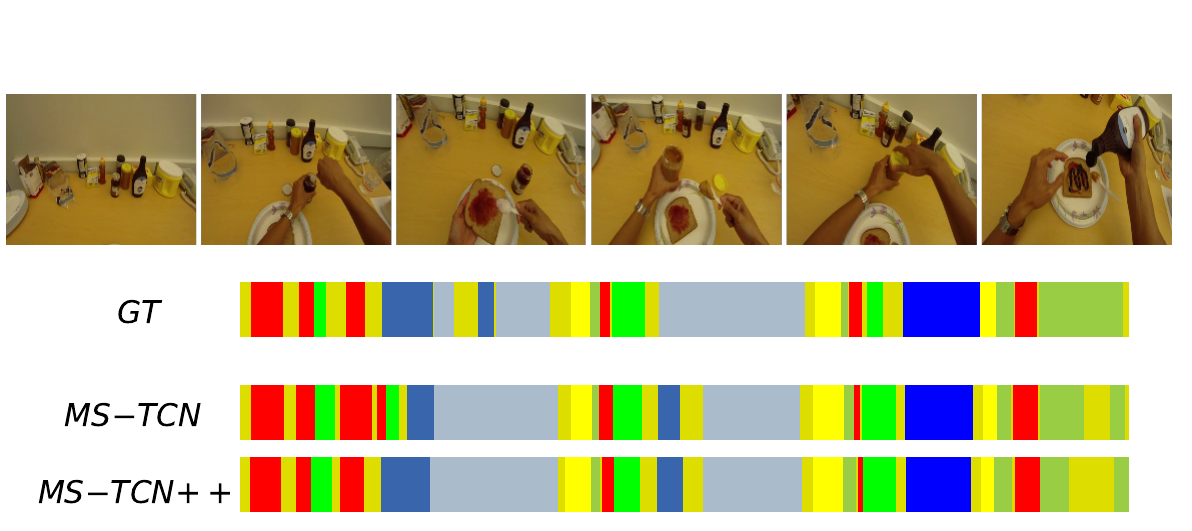}
}
\quad
\subfigure[]{
\includegraphics[trim={.5cm 4cm .5cm .5cm},clip,width=.47\linewidth]{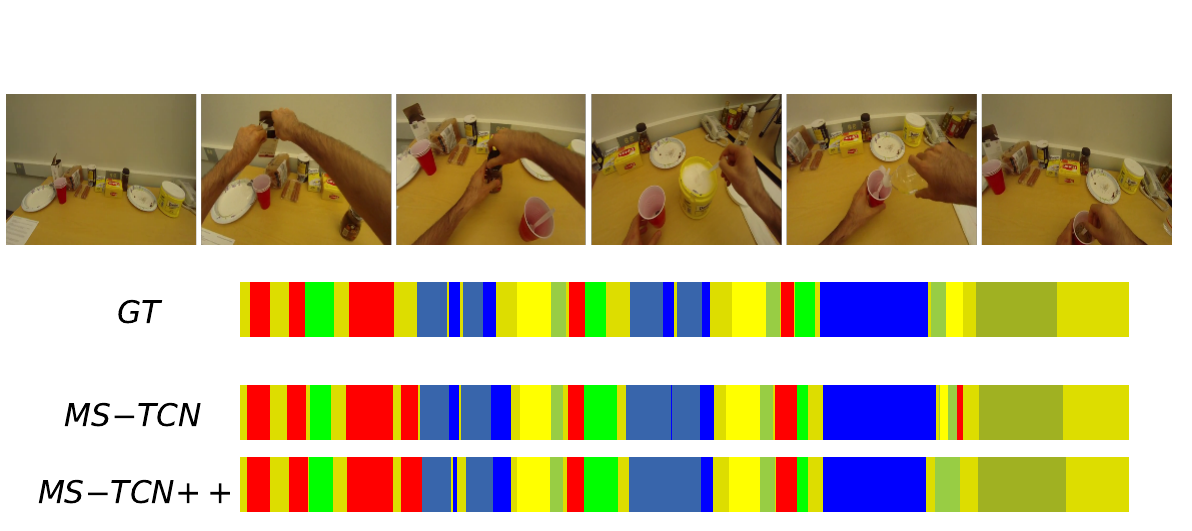}
}
\quad
\subfigure[]{
\includegraphics[trim={.5cm 4cm .5cm .5cm},clip,width=.47\linewidth]{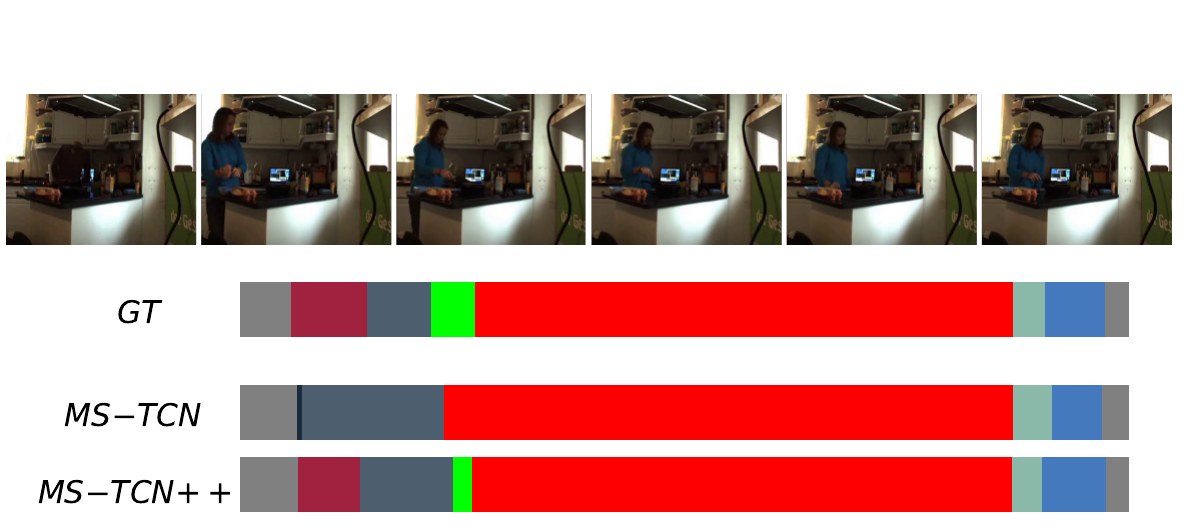}
}
\quad
\subfigure[]{
\includegraphics[trim={.5cm 4cm .5cm .5cm},clip,width=.47\linewidth]{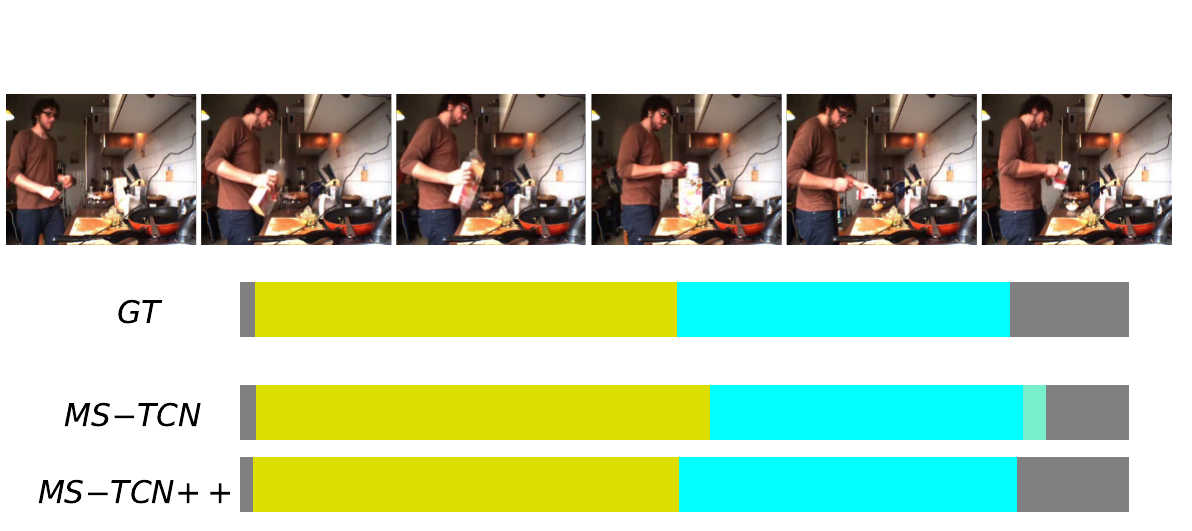}
}
\caption{Qualitative results for the temporal action segmentation task on (a)(b) 50Salads, (c)(d) GTEA, and (e)(f) Breakfast dataset.}
\label{fig:qualitative_res}
\end{figure*}


\subsection{Effect of the Number of Refinement Stages}
We set the number of refinement stages $N_r$ in MS-TCN++ to 3 stages, which results in a model with 4 stages 
in total. Table~\ref{tab:ref_stages} shows the impact of the refinement stages on the 50Salads dataset. Using 
only the prediction generation stage ($N_r = 0$) results in a relative low performance but it is much better 
than a single stage TCN (Table~\ref{tab:number_of_stages}). Adding more refinement stages 
improves the performance incrementally. Nevertheless, adding more than 3 refinement stages does not provide 
additional improvements. 

\begin{table}[tb]
\centering
\resizebox{\linewidth}{!}{%
\begin{tabular}{lcccccc}
\toprule
  & $N_r$ & \multicolumn{3}{c}{F1@\{10,25,50\}} & Edit & Acc  
\\ \midrule
MS-TCN++  & 0 	 &        51.0 &         48.4  &          40.7   &          40.4  &         80.7 \\
MS-TCN++  & 1    &        70.7 &         68.2  &          59.7   &          62.0  &         82.4 \\
MS-TCN++  & 2    &        77.8 &         75.1  &          66.9   &          69.4  &         82.5 \\
MS-TCN++  & 3    &\textbf{80.7} &        78.5  & \textbf{70.1}  &  \textbf{74.3} & \textbf{83.7}   \\
MS-TCN++  & 4    &        80.6 & \textbf{78.7} & \textbf{70.1}  &          73.1  &         82.4 \\
\bottomrule
\end{tabular}%
}
\caption{Impact of the number of refinement stages on the 50Salads dataset.}
\label{tab:ref_stages}
\end{table}


\subsection{Impact of Parameters Sharing}
MS-TCN++ consists of a prediction generation stage and 3 refinement stages. Although adding 
more stages results in a better performance, it also increases the number of parameters. 
As the refinement stages share in principle the same task, it is hence intuitive that they can share 
parameters. Table~\ref{tab:sharing} shows the impact of sharing parameters between the refinement 
stages. Sharing parameters significantly reduces the number of parameters with only a slight  
decrease in performance. For an MS-TCN++ with 3 refinement stages, sharing parameters reduces 
the total number of parameters to roughly $66\%$ of the total parameters in the original model. 
As shown in the table, MS-TCN++ with shared parameters outperforms MS-TCN with a margin of up to 
$3.8\%$ despite of having less parameters.

\begin{table}[tb]
\centering
\resizebox{\linewidth}{!}{%
\begin{tabular}{lccccccc}
\toprule
   & \multicolumn{3}{c}{F1@\{10,25,50\}} & Edit & Acc   & \# param.(m)
\\ \midrule
MS-TCN      	&        76.3 &        74.0 &    64.5     &       67.9  &       80.7              & 0.80  \\
MS-TCN++       	 &\textbf{80.7} & \textbf{78.5} & \textbf{70.1}  &  \textbf{74.3} & \textbf{83.7} & 0.99  \\
MS-TCN++(sh)     &        78.7 &         76.6  &          68.3   &          70.7  &         82.2  & 0.66\\
\bottomrule
\end{tabular}%
}
\caption{Impact of sharing parameters for the refinement stages on the 50Salads dataset.}
\label{tab:sharing}
\end{table}

\subsection{Impact of Temporal Resolution}
Previous temporal models operate on a low temporal resolution of 1-3 frames per second~\cite{lea2017temporal, lei2018temporal, ding2018weakly}. 
On the contrary, our approach is able to handle a higher resolution of 15 fps. In this experiment, we 
evaluate MS-TCN and MS-TCN++, with and without parameter sharing, for a low temporal resolution of 1 fps. As shown in Table~\ref{tab:temporal_res}, both models are able to handle both low and high temporal resolutions. 
While reducing the temporal resolution for MS-TCN results in a better edit distance and segmental F1 score, using a higher resolution gives better frame-wise accuracy. 
Operating on a low temporal resolution makes MS-TCN less prone to the over-segmentation problem, which is reflected in the better edit and F1 scores. 
Due to the dual dilated layers, MS-TCN++ benefits more from a higher temporal 
resolution and the impact of reducing the temporal resolution for MS-TCN++ is 
noticeable for all evaluation metrics. Note that 
even when we share the parameters of the refinement stages in MS-TCN++, using a higher temporal resolution 
results in a better performance.

\begin{table}[tb]
\centering
\resizebox{\linewidth}{!}{%
\begin{tabular}{lccccc}
\toprule
  & \multicolumn{3}{c}{F1@\{10,25,50\}} & Edit & Acc  
\\ \midrule
MS-TCN (1 fps)     & \textbf{77.8} & \textbf{74.9} &         64.0  & \textbf{70.7} &         78.6  \\
MS-TCN (15 fps)    &         76.3  &         74.0  & \textbf{64.5} &         67.9  & \textbf{80.7} \\ 
\midrule
MS-TCN++ (1 fps)     &        80.4 &  \textbf{78.7} &         68.6   &          73.3 &          81.1  \\
MS-TCN++ (15 fps)    &\textbf{80.7} &         78.5  & \textbf{70.1}  &  \textbf{74.3} & \textbf{83.7}   \\
\midrule
MS-TCN++(sh) (1 fps)     &        77.0 &         73.8  &          64.0   &          69.1  &         80.8  \\
MS-TCN++(sh) (15 fps)    & \textbf{78.7} & \textbf{76.6}  & \textbf{68.3}  & \textbf{70.7} &\textbf{82.2} \\
\bottomrule
\end{tabular}%
}
\vspace{1mm}
\caption{Impact of temporal resolution on the 50Salads dataset.}
\label{tab:temporal_res}
\end{table}


\subsection{Impact of Fine-tuning the Features}
In our experiments, we use the I3D features without fine-tuning. Table~\ref{tab:fine_tuning} 
shows the effect of fine-tuning on the GTEA dataset. Both of our multi-stage architectures, 
MS-TCN and MS-TCN++, significantly outperform the single stage architecture - with and without 
fine-tuning. This also holds when the parameters of the refinement stages in MS-TCN++ are shared. 
Fine-tuning improves the results, but the effect of fine-tuning for action segmentation 
is lower than for action recognition. This is expected since the temporal model is by far more important 
for segmentation than for recognition.

Note that without fine-tuning, sharing parameters achieves better results on GTEA. This is mainly due to the 
reduced number of parameters, which prevents the model from over-fitting the training data, especially for 
small datasets like GTEA. 

\begin{table}[ht]
\centering
\resizebox{\linewidth}{!}{%
\begin{tabular}{llccccc}
\toprule
&  & \multicolumn{3}{c}{F1@\{10,25,50\}} & Edit & Acc  
\\ \midrule
w/o FT   & SS-TCN            &         62.8  &         60.0  &         48.1  &         55.0  &         73.3  \\
         & MS-TCN            &         85.8  &         83.4  &         69.8  &         79.0  &         76.3  \\ 
         & MS-TCN++          &         87.0  &         85.2  &         73.5  &         82.0  &         78.7   \\
         & MS-TCN++(sh)      & \textbf{87.8} & \textbf{86.2} & \textbf{74.4} & \textbf{82.6} & \textbf{78.9}    \\
\midrule
with FT  & SS-TCN            &         69.5  &         64.9  &         55.8  &         61.1  &         75.3   \\
         & MS-TCN            &         87.5  &         85.4  &         74.6  &         81.4  &         79.2   \\
         & MS-TCN++          & \textbf{88.8} &         85.7  & \textbf{76.0} & \textbf{83.5} & \textbf{80.1}    \\
         & MS-TCN++(sh)      &         88.2  & \textbf{86.2}  &        75.9  &         83.0  &         79.7   \\
\bottomrule
\end{tabular}%
}
\vspace{1mm}
\caption{Effect of fine-tuning on the GTEA dataset.}
\label{tab:fine_tuning}
\end{table}


\begin{table}[tb]
\centering
\resizebox{\linewidth}{!}{%
\begin{tabular}{lccccc}
\toprule
\textbf{50Salads} & \multicolumn{3}{c}{F1@\{10,25,50\}} & Edit & Acc  
\\ \midrule
Spatial CNN\cite{lea2016segmental}  &        32.3  &        27.1  &        18.9   &        24.8  &        54.9 \\
IDT+LM\cite{richard2016temporal}    &        44.4  &        38.9  &        27.8   &        45.8  &        48.7 \\
Bi-LSTM\cite{singh2016multi}        &        62.6  &        58.3  &        47.0   &        55.6  &        55.7 \\
Dilated TCN\cite{lea2017temporal}   &        52.2  &        47.6  &        37.4   &        43.1  &        59.3 \\
ST-CNN\cite{lea2016segmental}       &        55.9  &        49.6  &        37.1   &        45.9  &        59.4 \\
TUnet\cite{ronneberger2015u}        &        59.3  &        55.6  &        44.8   &        50.6  &        60.6 \\
ED-TCN\cite{lea2017temporal}        &        68.0  &        63.9  &        52.6   &        52.6  &        64.7 \\
TResNet\cite{he2016deep}            &        69.2  &        65.0  &        54.4   &        60.5  &        66.0 \\
TRN\cite{lei2018temporal}           &        70.2  &        65.4  &        56.3   &        63.7  &        66.9 \\
TDRN+UNet\cite{lei2018temporal}     &        69.6  &        65.0  &        53.6   &        62.2  &        66.1 \\
TDRN\cite{lei2018temporal}          &        72.9  &        68.5  &        57.2   &        66.0  &        68.1 \\
LCDC+ED-TCN\cite{mac2019learning}   &        73.8  &          -   &          -    &        66.9  &        72.1 \\
\midrule
MS-TCN~\cite{abufarha2019tcn}       &        76.3  &        74.0  &        64.5   &        67.9  &        80.7 \\   
MS-TCN++(sh)                        &        78.7  &        76.6  &        68.3   &        70.7  &        82.2 \\
MS-TCN++                            &\textbf{80.7} &\textbf{78.5} &\textbf{70.1}  &\textbf{74.3} &\textbf{83.7}   \\
\bottomrule
\toprule
\textbf{GTEA} & \multicolumn{3}{c}{F1@\{10,25,50\}} & Edit & Acc  
\\ \midrule

Spatial CNN\cite{lea2016segmental}  & 41.8 & 36.0 & 25.1 &  -   & 54.1 \\
Bi-LSTM\cite{singh2016multi}        & 66.5 & 59.0 & 43.6 &  -   & 55.5 \\
Dilated TCN\cite{lea2017temporal}   & 58.8 & 52.2 & 42.2 &  -   & 58.3 \\
ST-CNN\cite{lea2016segmental}       & 58.7 & 54.4 & 41.9 &  -   & 60.6 \\
TUnet\cite{ronneberger2015u}        & 67.1 & 63.7 & 51.9 & 60.3 & 59.9 \\
ED-TCN\cite{lea2017temporal}        & 72.2 & 69.3 & 56.0 &  -   & 64.0 \\
LCDC+ED-TCN\cite{mac2019learning}   & 75.4 &   -  &   -  & 72.8 & 65.3 \\
TResNet\cite{he2016deep}            & 74.1 & 69.9 & 57.6 & 64.4 & 65.8 \\
TRN\cite{lei2018temporal}           & 77.4 & 71.3 & 59.1 & 72.2 & 67.8 \\
TDRN+UNet\cite{lei2018temporal}     & 78.1 & 73.8 & 62.2 & 73.7 & 69.3 \\
TDRN\cite{lei2018temporal}          & 79.2 & 74.4 & 62.7 & 74.1 & 70.1 \\
\midrule
MS-TCN~\cite{abufarha2019tcn}  &        87.5  &        85.4  &        74.6  &        81.4  &        79.2 \\ 
MS-TCN++(sh)                   &        88.2  &\textbf{86.2} &        75.9  &        83.0  &        79.7   \\
MS-TCN++                       &\textbf{88.8} &        85.7  &\textbf{76.0} &\textbf{83.5} &\textbf{80.1}   \\
\bottomrule
\toprule
\textbf{Breakfast} & \multicolumn{3}{c}{F1@\{10,25,50\}} & Edit & Acc  
\\ \midrule
ED-TCN~\cite{lea2017temporal}*  &          -    &          -    &          -    &          -    &         43.3  \\
HTK~\cite{kuehne2017weakly}   &          -    &          -    &          -    &          -    &         50.7  \\
TCFPN~\cite{ding2018weakly}   &          -    &          -    &          -    &          -    &         52.0  \\
HTK(64)~\cite{kuehne2016end}  &          -    &          -    &          -    &          -    &         56.3  \\   
GRU~\cite{richard2017weakly}* &         -    &          -    &          -    &          -    &         60.6  \\ 
GRU+length prior~\cite{kuehne2018hybrid} &   -    &  -    &          -    &          -    &         61.3  \\ 
\midrule
MS-TCN (IDT)~\cite{abufarha2019tcn}     &        58.2   &         52.9  &       40.8   &      61.4  &         65.1  \\      
MS-TCN (I3D)~\cite{abufarha2019tcn}     &        52.6   &         48.1  &       37.9   &      61.7  &         66.3 \\      
MS-TCN++(I3D) (sh)                      &         63.3  & 57.7 &       44.5   & 64.9 &       67.3 \\
MS-TCN++ (I3D)                          & \textbf{64.1} &         \textbf{58.6}  &\textbf{45.9} &        \textbf{65.6} & \textbf{67.6} \\
\bottomrule
\end{tabular}%
}
\vspace{1mm}
\caption{Comparison with the state-of-the-art on 50Salads, GTEA, and the Breakfast dataset. (* obtained from~\cite{ding2018weakly}).}
\label{tab:state_of_the_art}
\vspace{-4mm}
\end{table}

\subsection{Comparison with the State-of-the-Art}
In this section, we compare the proposed models to the state-of-the-art methods on three datasets: 50Salads, Georgia Tech Egocentric Activities (GTEA), and the Breakfast datasets. 
The results are presented in Table~\ref{tab:state_of_the_art}. 
As shown in the table, our models outperform the state-of-the-art methods on the three datasets and with respect to three evaluation metrics: F1 score, segmental edit distance, and frame-wise accuracy (Acc) with a large margin (up to $11.6 \%$ for the frame-wise accuracy on the 50Salads dataset). 
Qualitative results on the three datasets are shown in Figure~\ref{fig:qualitative_res}. 
Note that all the reported results are obtained using the I3D features. 
To analyze the effect of using a different type of features, we evaluated MS-TCN on the Breakfast dataset using the improved dense trajectories (IDT) features, which are the commonly used features for the Breakfast dataset. 
As shown in Table~\ref{tab:state_of_the_art}, the impact of the features is very small. 
While the frame-wise accuracy and edit distance are slightly better using the I3D features, the model achieves a better F1 score when using the IDT features compared to I3D. 
This is mainly because I3D features encode both motion and appearance, whereas the IDT features encode only motion. 
For datasets like Breakfast, using appearance information does not help the performance since the appearance does not give a strong evidence about the action that is carried out. 
This can be seen in the qualitative results shown in 
Figure~\ref{fig:qualitative_res}. 
The video frames share a very similar appearance. 
Additional appearance features therefore do not help in recognizing the activity.
As shown in Table~\ref{tab:state_of_the_art}, sharing the parameters of the refinement 
stages achieves similar performance to MS-TCN++, but it requires about $66\%$ less parameters as reported in Table~\ref{tab:sharing}.

As our approaches do not use any recurrent layers, they are very fast both during training and testing. 
Training MS-TCN++ for $50$ epochs takes only $10$ minutes on the 50Salads dataset compared to $35$ minutes 
for training a single cell of a Bi-LSTM with a 64-dimensional hidden state on a single GTX 1080 Ti GPU. 
This is due to the sequential prediction of the LSTM, where the activations at any time step depend on the activations from the previous steps. 
For the MS-TCN and MS-TCN++, activations at all time steps are computed in parallel.

\section{Conclusion}
We presented two multi-stage architectures for the temporal action segmentation task. 
While the first stage generates an initial prediction, this prediction is iteratively 
refined by the higher stages. 
Instead of the commonly used temporal pooling, we used dilated convolutions to increase 
the temporal receptive field. The experimental evaluation demonstrated the capability of 
our architecture in capturing temporal dependencies between action classes and reducing 
over-segmentation errors. We further introduced a smoothing loss that gives an additional 
improvement of the predictions quality. We also introduced a dual dilated layer that captures 
both local and global features, which results in better performance. Moreover, we showed that 
sharing the parameters in the refinement stages results in a more efficient model with a slight 
degradation in performance. Our models outperform the state-of-the-art methods on three 
challenging datasets recorded from different views with a large margin. Since our model is 
fully convolutional, it is very efficient and fast both during training and testing.


%



\ifCLASSOPTIONcompsoc
  \section*{Acknowledgments}
\else
  \section*{Acknowledgment}
\fi

The work has been funded by the Deutsche Forschungsgemeinschaft (DFG, German Research Foundation) – GA 1927/4-1 (FOR 2535 Anticipating Human Behavior) and the ERC Starting Grant ARCA (677650).

\ifCLASSOPTIONcaptionsoff
  \newpage
\fi



%

\bibliographystyle{IEEEtran}
\bibliography{egbib}

%
%

%

\begin{IEEEbiography}[{\includegraphics[width=1in,height=1.25in,clip,keepaspectratio]{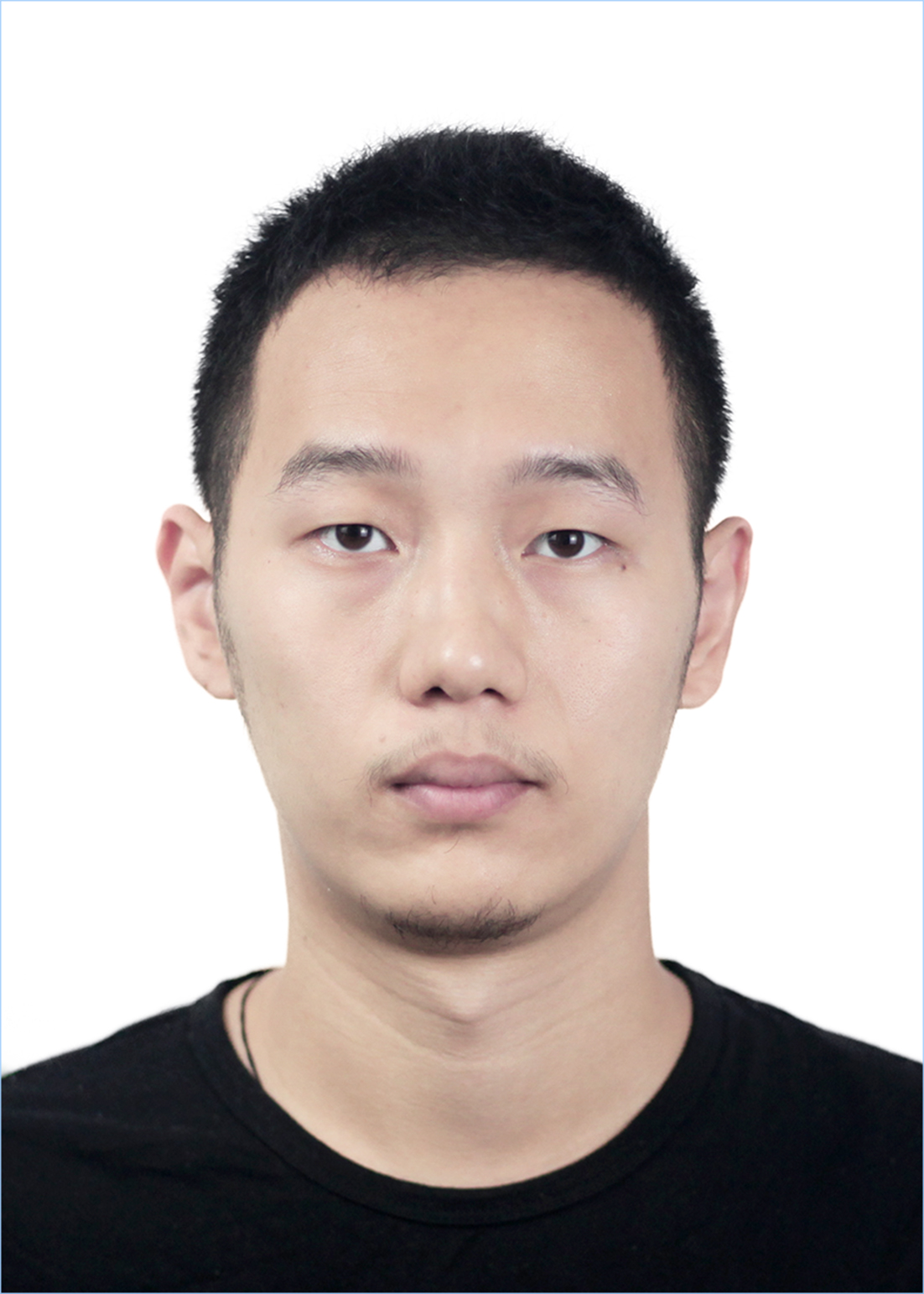}}]{Shijie Li}
received his Bachelor degree in Automation Engineering from University of Electronic Science and Technology of China in 2016 and his Master degree in computer science from the Nankai University in 2019. Since 2019, he is a PhD student at the University of Bonn. His research interests include action recognition and scene understanding.
\end{IEEEbiography}

\begin{IEEEbiography}[{\includegraphics[width=1in,height=1.25in,clip,keepaspectratio]{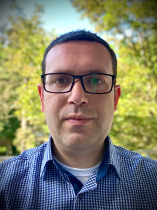}}]{Yazan Abu Farha}
received his Bachelor degree in computer systems engineering from Birzeit University in 2013 and his Master degree in computer science from the University of Bonn in 2017. Since 2018, he is a PhD student at the University of Bonn. His research interests include action recognition and anticipation.
\end{IEEEbiography}

\begin{IEEEbiography}[{\includegraphics[width=1in,height=1.25in,clip,keepaspectratio]{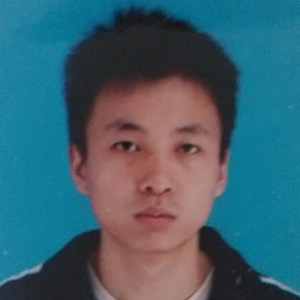}}]{Yun Liu}
is a PhD candidate at College of Computer Science, Nankai University. He received
his bachelor degree from Nankai University in
2016. His research interests include computer
vision and machine learning.
\end{IEEEbiography}

\begin{IEEEbiography}[{\includegraphics[width=1in,height=1.25in,clip,keepaspectratio]{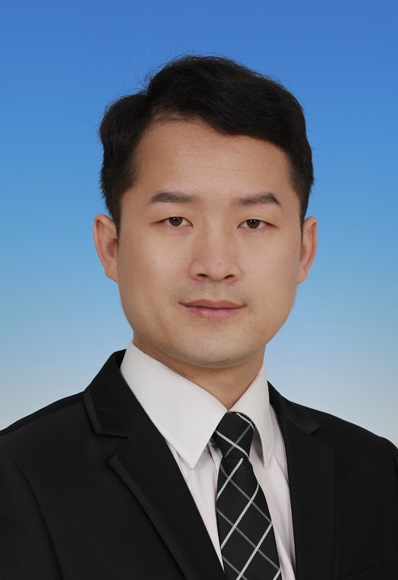}}]{Ming-Ming Cheng}
received his PhD degree
from Tsinghua University in 2012. Then he did 2
years research fellow, with Prof. Philip Torr in Oxford. He is now a professor at Nankai University,
leading the Media Computing Lab. His research
interests includes computer graphics, computer
vision, and image processing. He received research awards including ACM China Rising Star
Award, IBM Global SUR Award, and CCF-Intel
Young Faculty Researcher Program. He is on the
editorial boards of IEEE TIP.
\end{IEEEbiography}

\begin{IEEEbiography}[{\includegraphics[width=1in,height=1.25in,clip,keepaspectratio]{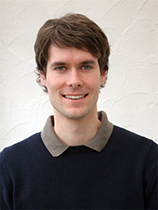}}]{Juergen Gall}

obtained his B.Sc.
and his Masters degree in mathematics from the
University of Wales Swansea (2004) and from
the University of Mannheim (2005). In 2009, he
obtained a Ph.D. in computer science from the
Saarland University and the Max Planck Institut für Informatik. He was a postdoctoral researcher at the Computer Vision Laboratory, ETH Zurich,
from 2009 until 2012 and senior research scientist at the Max Planck Institute for Intelligent Systems in Tübingen from 2012 until 2013. Since 2013, he is professor at the University of Bonn and head of the Computer
Vision Group.

\end{IEEEbiography}







\end{document}